\documentclass[twoside]{article}
\usepackage[preprint]{aistats2027}

\usepackage[round]{natbib}

\usepackage{amsmath,amssymb,amsthm}
\usepackage{bm}
\usepackage{booktabs}
\usepackage{multirow}
\usepackage{float}
\usepackage{graphicx}
\usepackage{caption}
\usepackage{enumitem}
\usepackage{xcolor}
\usepackage{hyperref}
\usepackage{url}
\usepackage{subcaption}
\newcommand{\sv}{\mathbf{s}}

\graphicspath{{figures/}}
\newtheorem{theorem}{Theorem}[section]
\newtheorem{lemma}[theorem]{Lemma}
\newtheorem{definition}[theorem]{Definition}
\newtheorem{proposition}[theorem]{Proposition}
\newtheorem{corollary}[theorem]{Corollary}
\theoremstyle{remark}
\newtheorem{remark}[theorem]{Remark}
\newtheorem{assumption}[theorem]{Assumption}
\newtheorem{conjecture}[theorem]{Conjecture}

\def\real{\mathbb R}

\newcommand{\inner}[2]{\langle #1, #2 \rangle}
\DeclareMathOperator{\rank}{rank}
\def\sms{\mathrm{sp}}
\def\nuc{\mathrm{nuc}}
\def\msp{\mathrm{msp}}
\def\Cc{C^{\circ}}  
\providecommand{\R}{\mathbb{R}}

\def\T{^{\top}}

\newcommand{\x}{\mathbf{x}}
\newcommand{\thv}{\bm{\theta}}
\newcommand{\uu}{\mathbf{u}}
\newcommand{\vv}{\mathbf{v}}
\newcommand{\pp}{\mathbf{p}}
\newcommand{\qq}{\mathbf{q}}
\newcommand{\ee}{\mathbf{e}}
\newcommand{\sgn}{\operatorname{sign}}
\newcommand{\tr}{\operatorname{tr}}
\newcommand{\dstar}{\star}

\newcommand{\kk}{\mathbf{k}}

\newcommand{\bx}{\mathbf{x}}

\newcommand{\btheta}{\boldsymbol{\theta}}

\DeclareMathOperator{\conv}{conv}

\makeatletter
\@ifundefined{conv}{\DeclareMathOperator{\conv}{conv}}{}
\@ifundefined{cone}{\DeclareMathOperator{\cone}{cone}}{}
\@ifundefined{aff}{\DeclareMathOperator{\aff}{aff}}{}
\@ifundefined{supp}{}{}
\@ifundefined{argmin}{}{}
\@ifundefined{argmax}{\DeclareMathOperator*{\argmax}{arg\,max}}{}
\@ifundefined{codim}{\DeclareMathOperator{\codim}{codim}}{}
\makeatother

\newcommand{\bk}{{\bf k}}

\newcommand{\br}{{\bf r}}
\newcommand{\rr}{{\bf r}}
\usepackage{mathtools}

\newcommand{\B}{{\mathcal{B}}}

\begin{document}

\twocolumn[

\aistatstitle{Which Optimizer Leaks Most? Optimizer Implicit Bias as a Training-Data Privacy Controller}

\aistatsauthor{ Sagi Habani$^*$ \And Yehonathan Refael$^*$ \And  Ofir Lindenbaum }

\aistatsaddress{ Ben-Gurion University \\ sagihab@post.bgu.ac.il \And  Tel Aviv University \\  refaelkalim@mail.tau.ac.il \And Bar Ilan University \\ ofir.lindenbaum@biu.ac.il } ]

\begin{abstract}
Releasing a neural network checkpoint can leak information about its training data, and existing reconstruction results rely on gradient-descent training \citep{haim2022}. We ask whether the choice of optimizer affects the ability to recover the training data. Extending work connecting Karush--Kuhn--Tucker conditions to training-data reconstruction, we show that the implicit biases of popular optimizers (SGD, Adam, Muon) leave very different fingerprints of the training data. We argue that the key distinction is the \emph{weight-specificity} of these implicit biases, namely how much continuous information about the final weights the optimizer-dependent target of the stationarity conditions encodes. The limiting stationarity conditions for models trained with SGD and Muon retain enough weight-specific information to reconstruct training examples in practice. For Adam in the $\ell_\infty$ implicit-bias regime, the optimizer-dependent part of these conditions encodes only a discrete pattern of maximal coordinates and their signs, and at nondegenerate KKT points the conditions are too few to identify the training set to first order. Experiments on MNIST and CIFAR-10 support these claims: models trained with Muon can yield stronger reconstructions at much lower training cost than models trained with SGD, although this advantage depends on the training regime and architecture, and models trained with Adam appear consistently more resistant to the evaluated attacks.
\end{abstract}

\begin{NoHyper}
  {\let\thefootnote\relax\footnotetext{$^*$Equal contribution.}}
\end{NoHyper}

\section{Introduction}
\label{sec:intro}
\begin{figure*}[htb]
  \centering
  \begin{subfigure}[b]{0.32\textwidth}
    \centering
    \includegraphics[width=\linewidth]{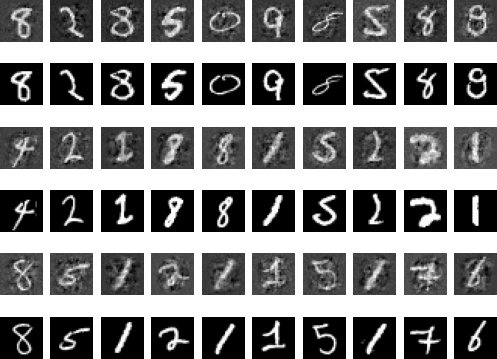}
    \caption{Muon}
    \label{fig:mnist_muon_noprior}
  \end{subfigure}\hfill
  \begin{subfigure}[b]{0.32\textwidth}
    \centering
    \includegraphics[width=\linewidth]{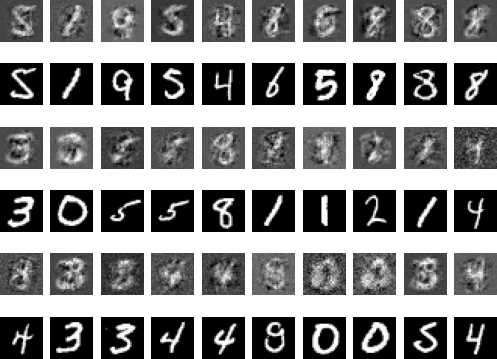}
    \caption{SGD}
    \label{fig:mnist_sgd_noprior}
  \end{subfigure}\hfill
  \begin{subfigure}[b]{0.32\textwidth}
    \centering
    \includegraphics[width=\linewidth]{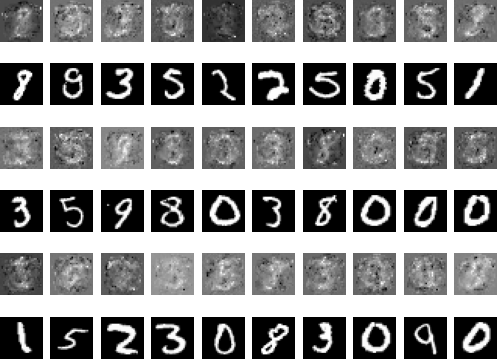}
    \caption{Adam}
    \label{fig:mnist_adam_noprior}
  \end{subfigure}
  
  \vspace{0.4em}
  \begin{subfigure}[b]{0.32\textwidth}
    \centering
    \includegraphics[width=\linewidth]{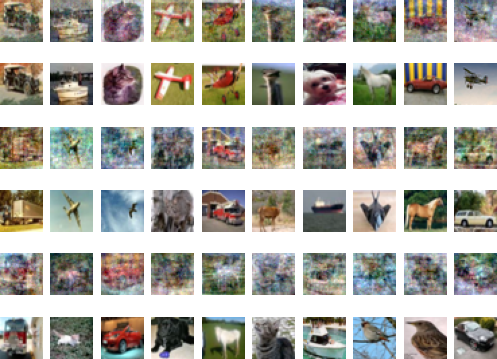}
    \caption{Muon}
    \label{fig:cifar10_muon_noprior}
  \end{subfigure}\hfill
  \begin{subfigure}[b]{0.32\textwidth}
    \centering
    \includegraphics[width=\linewidth]{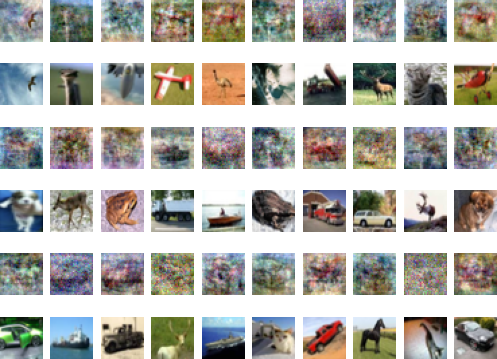}
    \caption{SGD}
    \label{fig:cifar10_sgd_noprior}
  \end{subfigure}\hfill
  \begin{subfigure}[b]{0.32\textwidth}
    \centering
    \includegraphics[width=\linewidth]{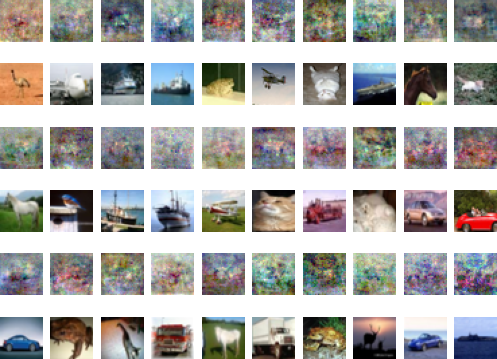}
    \caption{Adam}
    \label{fig:cifar10_adam_noprior}
  \end{subfigure}
  \caption{Training-data reconstructions from networks trained with Muon, SGD and Adam, without an
image prior ($n=500$). The top row is binary MNIST and the bottom row binary CIFAR-10. In each panel, odd rows show reconstructions and the row directly below
shows their nearest training images under the one-to-one matching of Section~\ref{sec:exp:setup}.
Muon reconstructions are visually closest to their matches, SGD recovers coarser structure, and
Adam reconstructions are largely unrecognizable.}
  \label{fig:optimizer_comparison}
\end{figure*}
 
Over-parameterized networks can perform well while memorizing training examples~\citep{zhang2017}, and good generalization may require memorizing atypical examples~\citep{feldman2020}, raising a privacy concern: how much memorized data can be recovered?
Recently, \citet{haim2022,buzaglo2023} demonstrated the ability to reconstruct a large fraction of the training images used to train a classifier, using only the parameters of the trained classifier (without query access or other data). They achieve this by optimizing the input to satisfy a stationarity condition that relates the weights of the neural network to the training data.
Merely sharing a trained model (a checkpoint) may therefore leak its training data.

This attack exploits the {\em implicit bias} of gradient descent (GD): for a homogeneous network trained with an exponentially tailed loss, the parameter direction converges, under suitable conditions, to that of a KKT point of the maximum-margin problem $\min_{\thv} \tfrac{1}{2}\|\thv\|_2^2, $ s.t.\ $y_i f(\x_i;\thv)\ge1$ for all $i$ \citep{Lyu2020Gradient,jitelgarsky2020}. Its stationarity condition expresses the weights as a nonnegative combination of the signed gradients $y_i \nabla_{\thv}f(\x_i;\thv)$, and~\citet{haim2022} invert it by solving for inputs $\x_i$ with the weights fixed: the optimizer's bias gives the attacker a system of equations relating the public weights to the training data.

Other optimizers induce other margin geometries: under suitable assumptions for smooth homogeneous models, Muon \citet{jordan2024muon} maximizes a max-spectral-norm margin and Adam~\citep{kingma2015adam}, with a negligible stability constant, an $\ell_\infty$ margin~\citep{gronich2026}. Different optimizers therefore induce different stationarity equations; we analyze how these differences affect data reconstruction and whether the choice of optimizer can reduce this risk.

We generalize the objective of \citet{haim2022} to the margin geometry of the chosen optimizer (Section~\ref{sec:obj}). A direct extension replaces its Euclidean stationarity term with a condition on the subdifferential of the margin norm. When applied naively, however, a formulation based on a single leading singular-vector pair may fail when the top singular value is repeated (a property encouraged by Muon) or when the relative block scales it fixes differ from those realized at a KKT point. It also breaks the rescaling symmetry of the KKT system, so it can be lowered by sending the multipliers to zero rather than by aligning the candidates with the data (Lemma~\ref{prop:scale}). To avoid this problem, we derive a scale-invariant penalty that enforces equality in the dual-norm inequality and use the layerwise Euler identity to show that no cross-block coupling term is needed.

Our analysis (Section~\ref{sec:theory}) identifies two properties that help explain why the attack works. The first is {\em weight-specificity} -- the number of independent continuous functionals of $\thv$ that the optimizer-dependent stationarity target determines. Equation counting at a generic weight vector does not distinguish the three geometries (there the target set has codimension $\Theta(p)$ in all three, where $p$ is the parameter count), but specificity does. At a KKT direction, where every block is active (Corollary~\ref{cor:active}), and with simple top singular values ($\rho_k=1$), $S$ has the values
$S_{\ell_2} = p - 1,\quad
S_{\text{spectral}} = \sum_{k} (m_k + n_k - 2),\quad
S_{\ell_\infty} = 0,$
where block $k$ has dimensions $m_k\times n_k$, so each spectral block contributes a number of functionals proportional to its width. For the $\ell_\infty$ margin, the optimizer-dependent target depends only on the locations and signs of the coordinates of $\thv$ with maximum absolute value, regardless of the multiplicity of the maximum. Moreover, at nondegenerate KKT points, the stationarity constraints impose at most as many equations as there are on-margin examples, far fewer than the unknowns, so the training set is not identifiable from them to first order.

The second property is {\em reachability} - the speed of convergence to the limiting parameter direction of the optimizer. Combining the two, we find the practical leakage ordering $\text{Muon}\succ\text{SGD}\succ\text{Adam}$ in our main experiments, where $\succ$ denotes greater reconstruction leakage (Section~\ref{sec:exp}). The Muon comparison depends on the training budget, sample size, and architecture (SGD becomes comparable to Muon at small sample sizes and matches or exceeds it in some architectures), while Adam remains the most resistant to the evaluated attacks; we do not, however, provide a formal privacy guarantee.

\paragraph{Our contributions are:} (i) a reconstruction attack that depends on the optimizer's margin norm, recovers the objective of \citet{haim2022} in the Euclidean case up to the multiplier scale, and corrects two failure modes of the naive extension (Section~\ref{sec:obj}); (ii) a theory of \emph{weight-specificity} under which Adam's $\ell_\infty$ target has zero specificity at every multiplicity of the maximum and its nondegenerate KKT points fail regular first-order identifiability (Section~\ref{sec:theory}); (iii) an analysis combining specificity and \emph{reachability} that explains the ordering $\text{Muon}\succ\text{SGD}\succ\text{Adam}$ in our main experiments, with experiments showing that the Muon comparison depends on the training regime while Adam remains the most resistant to the evaluated attacks (Section~\ref{sec:exp}).

\section{Related Work}
\label{sec:related}

\paragraph{Reconstruction among privacy attacks.} \citet{haim2022} reconstruct training samples from homogeneous classifiers by inverting the $\ell_2$-margin KKT conditions; follow-up work extends this to the multiclass setting, weight decay and general losses \citep{buzaglo2023}, to transfer learning \citep{oz2024}, and to membership inference \citep{golbari2025}, and examines the specific reliance of reconstruction attacks on inductive image priors \citep{refael2026no}. All use the $\ell_2$ implicit bias of gradient descent; we extend the formulation to other margin geometries and study how the optimizer affects the attack.

\paragraph{Implicit bias, and the geometry of Adam and Muon.} Adam has a GD-like implicit bias when its stability constant dominates \citep{wang2021} and an $\ell_\infty$ margin bias when it is absent \citep{zhang2024}; spectral descent and Muon have been analyzed in the multiclass linear setting \citep{fan2025}. Under suitable assumptions, \citet{gronich2026} extend these results to smooth homogeneous networks, relating idealized Muon to normalized momentum steepest descent under the \emph{max-spectral} norm and Adam to the \emph{global} $\ell_\infty$ margin, and leave open whether KKT-based reconstruction extends to these optimizers. We derive attacks for both geometries: reconstruction succeeds for Muon, while our analysis identifies a first-order identifiability limitation of Adam's stationarity constraints, consistent with its greater resistance in our experiments.

\subsection{Preliminaries: notation and threat model}
\label{sec:notation}

\paragraph{Notation.} We write $[n]:=\{1,\dots,n\}$, $(a)_+:=\max\{a,0\}$, $\ee_j$ for the $j$-th
standard basis vector, and $S^{d-1}$ for the unit sphere in $\R^d$. We denote
the convex, affine, and conic hulls of a set $X$ by $\conv X$, $\aff X$, and
$\cone X$, respectively. For a matrix $A$, $\sigma_1(A)\ge\sigma_2(A)\ge\cdots$
are its singular values, $A\succeq0$ means that $A$ is positive semidefinite,
and $\{M\succeq0:\tr M=1\}$ is a spectrahedron.
We write \(S^\circ:=S\setminus\{0\}\).

\paragraph{Parameters, blocks and norms.} The network has parameters $\thv=(W_1,\dots,W_K)$, where $W_k\in\R^{m_k\times n_k}$ and each weight matrix is a \emph{block}. We identify $\thv$ with the concatenated vector in $\R^p$, where $p=\sum_k m_kn_k$. We write $\theta_j$ for its $j$-th coordinate and $\uu_{W_k}$ for the $k$-th block of any $\uu\in\R^p$, and use the Euclidean inner product $\inner{\uu}{\vv}=\sum_k\tr(\uu_{W_k}\T\vv_{W_k})$. For each block, we use the Frobenius, spectral, and nuclear norms $\|W\|_F$, $\|W\|_{\sms}=\sigma_1(W)$, and $\|W\|_{\nuc}=\sum_i\sigma_i(W)$, respectively. On the full parameter vector, we use $\|\thv\|_1$, $\|\thv\|_2=(\sum_k\|W_k\|_F^2)^{1/2}$, $\|\thv\|_\infty=\max_j|\theta_j|$, and the \emph{max-spectral} norm $\|\thv\|_{\msp}=\max_k\|W_k\|_{\sms}$ of \citet{gronich2026}. An unadorned $\|\cdot\|$ denotes the \emph{margin norm} of the optimizer under discussion (Eq.~\eqref{eq:maximal_margin}), and $\|\cdot\|_\dstar$ denotes its \emph{dual norm},
$\|\uu\|_\dstar=\sup\{\inner{\uu}{\vv}:\|\vv\|\le1\}.$
These norms satisfy $\inner{\uu}{\vv}\le\|\uu\|_\dstar\|\vv\|$, which we refer to as the \emph{generalized Cauchy--Schwarz inequality}.

\paragraph{Alignment, gauge, strata.} We write $\partial g(\thv)$ for the subdifferential of a convex function $g$ at $\thv$. Where $g$ is differentiable, this is the singleton $\{\nabla g(\thv)\}$. For a norm,
$\partial\|\thv\|=\{\uu:\|\uu\|_\dstar\le1,\ \inner{\uu}{\thv}=\|\thv\|\}.$
We say that $\uu$ is \emph{dual-aligned} with $\thv$ when equality holds in generalized Cauchy--Schwarz:
\begin{equation}\label{eq:aligndef}
\begin{aligned}
    \inner{\uu}{\thv} &= \|\uu\|_\dstar\|\thv\|, \\
    \text{equivalently } & \uu/\|\uu\|_\dstar \in \partial\|\thv\| \quad (\uu \neq \mathbf{0}).
\end{aligned}
\end{equation}

A \emph{gauge} is a group of transformations of the unknowns that preserves the constraint system and maps solutions to solutions. Since the constraints cannot distinguish such solutions, we make our objective invariant along gauge directions.

Finally, we divide parameter space into \emph{strata} on which the structural type of $\partial\|\thv\|$ is constant (Appendix~\ref{app:notation}).

\paragraph{Blockwise homogeneity.} Our structural results use homogeneity layer by layer.

\begin{assumption}[Blockwise homogeneity]\label{ass:blockhom}
For each block $k$ there is $d_k>0$ such that $f(\x;W_1,\dots,\alpha W_k,\dots,W_K)=\alpha^{d_k}f(\x;\thv)$
for all $\alpha>0$, and $f(\x;\cdot)$ is differentiable at the points considered. Then $f$ is
$L$-homogeneous with $L=\sum_kd_k$.
\end{assumption}

\paragraph{Optimizer implicit bias.} Homogeneous networks trained with logistic or exponential loss converge in direction to a KKT point of a maximum-margin problem; this implicit bias is the key phenomenon underlying the attacks we study. For gradient flow, this follows from  Theorem~\ref{thm:implicit_bias} in Appendix~\ref{app:implicit-bias}. Under the assumptions of  \citet{gronich2026}, Adam and Muon converge to KKT directions of (P), with the optimizer-dependent margin norm given in (2).
\begin{equation}\label{eq:maximal_margin}
\min_{\thv}\ \tfrac12\|\thv\|^2 \quad\text{s.t.}\quad y_i \Phi(\x_i;\thv)\ge 1\ \ \forall i\in[n].
\tag{P}
\end{equation}

Since every count below depends on how that norm aggregates across blocks, we fix it explicitly:
\begin{equation}\label{eq:norms}
\begin{aligned}
    \|\thv\|_{\mathrm{SGD}}  &= \Big(\textstyle\sum_k\|W_k\|_F^2\Big)^{1/2}, \\
    \|\thv\|_{\mathrm{Muon}} &= \max_k \frac{\|W_k\|_{\sms}}{c_k}, \\
    \|\thv\|_{\mathrm{Adam}} &= \|\thv\|_\infty.
\end{aligned}
\end{equation}
with duals $\ell_2$, $\sum_k c_k\|\cdot\|_{\nuc}$ and $\ell_1$. For $c_k\equiv1$, the Muon form is
$\|\cdot\|_{\msp}$ used by \citet{gronich2026} (their Corollary~3.3); general $c_k>0$ are the per-block
weights induced by practical Muon's per-block step sizes (their Theorem~3.6). All our statements for
Muon hold for arbitrary $c_k$, and the objective we construct does not depend on them. The Adam
$\ell_\infty$ norm is global over all $p$ coordinates rather than per tensor. This margin norm is the
\emph{only} property of the optimizer the attack uses, and the only axis along which our three
optimizers differ in the analysis.

\paragraph{Threat model.} Following \citet{haim2022}, the adversary observes the final parameters
$\thv$ and the architecture and knows the optimizer \emph{family}, hence $\|\cdot\|$; it observes
neither the training inputs nor their labels or count $n$, nor any intermediate iterate. In particular
$\thv$ is \emph{fixed and given}: the attacker may not rescale it (Section~\ref{sec:obj}). As is standard, the attacker posits $m$ candidates (we use $m=2n$ following \citet{haim2022}, as an upper bound on $n$ suffices) with fixed balanced labels and recovers a subset corresponding to true samples.

\paragraph{The gradient-descent attack \citep{haim2022}.} For $\ell_2$,
$\partial(\tfrac12\|\thv\|_2^2)=\{\thv\}$, so stationarity reads
$\thv=\sum_i\lambda_i y_i\nabla_{\thv} \Phi(\x_i;\thv)$ with $\lambda_i\ge0$, and complementary slackness
forces $\lambda_i=0$ off the margin. With $\thv$ known and the inputs unknown, \citet{haim2022} minimize
over candidate inputs and multipliers
\begin{equation}\label{eq:haim}
\begin{aligned}
    L_{\mathrm{rec}} &= \underbrace{\Big\|\thv-\textstyle\sum_i\lambda_i y_i \nabla_{\thv} \Phi(\x_i;\thv)\Big\|_2^2}_{L_{\mathrm{stat}}} \\
    &\quad + a_2\underbrace{\textstyle\sum_i(-\lambda_i)_+}_{L_\lambda} + a_3\, L_{\mathrm{prior}},
\end{aligned}
\end{equation}
where $L_\lambda$ enforces $\lambda_i\ge0$ and $L_{\mathrm{prior}}$ is a weak input prior such as a pixel-range penalty. Only $L_{\mathrm{stat}}$ depends on the optimizer, and only through
$\partial(\tfrac12\|\thv\|^2)$. That is the starting point of the next section.

\section{A Norm-Parameterized Reconstruction Objective}
\label{sec:obj}

We build the attack in three steps: the stationarity condition of~\eqref{eq:maximal_margin} for a general margin norm (\S\ref{sec:obj:stat}); removal of a non-informative degree of freedom, which yields an alignment condition and our loss (\S\ref{sec:obj:gauge}); and its instantiation for Muon (\S\ref{sec:obj:muon}) and Adam (\S\ref{sec:obj:adam}). Throughout, the trained $\thv$ is fixed, the unknowns are $z := (\{ \x_i \}_{i=1}^m, \{ \lambda_i \}_{i=1}^m)$, and the residual is
\begin{equation}
r(z) := \sum_{i=1}^m \lambda_i y_i \nabla_{\thv} \Phi(\x_i; \thv).
\end{equation}

\subsection{Stationarity for a general norm}
\label{sec:obj:stat}

For (S)GD the margin norm is Euclidean, \(\tfrac12 \|\thv\|_2^2\) has gradient \(\thv\), and the stationarity term of \eqref{eq:haim} vanishes iff \(\br = \thv\). For Muon and Adam, \(\tfrac12 \|\thv\|^2\) is not differentiable, so we replace the gradient with the subdifferential; by the chain rule \(\partial (\tfrac12 \|\thv\|^2) = \|\thv\| \partial \|\thv\|\) (Lemma~\ref{lem:half}), and the stationarity condition for \eqref{eq:maximal_margin} is
\begin{equation}
\label{eq:member}
\br(z) \in \partial \bigl(\tfrac12 \|\thv\|^2\bigr), \qquad \lambda_i \ge 0.
\end{equation}
To obtain a tractable loss, the following lemma characterizes this set-membership condition in terms of two scalar inequalities.
\begin{lemma}[Membership rule]
\label{prop:member}
For \(\theta\neq0\), \(\br\in\partial\bigl(\tfrac12\|\thv\|^2\bigr)\) if and only if
\(\|\br\|_\dstar\le\|\thv\|\) and
\(\inner{\br}{\thv}\ge\|\thv\|^2\).
\end{lemma}
Lemma~\ref{prop:member} splits \eqref{eq:member} into a \emph{magnitude} and an \emph{alignment} condition (generalizing \eqref{eq:aligndef}). Both \(\|\br(z)\|_{\dstar}\) and \(\inner{\br(z)}{\thv}\) are computable and almost everywhere differentiable in \(z\), and both inequalities hold with equality for feasible \(\br\) (see the proof), so \eqref{eq:member} can be enforced by penalizing \(\|\br\|_\dstar = \|\thv\|\) and \(\inner{\br}{\thv} = \|\thv\|^2\). As we show next, all information about \(\{\x_i\}\) lies in the \emph{alignment} requirement.

\subsection{Two scale freedoms, and why only one may be discarded}
\label{sec:obj:gauge}

The magnitude bound of Lemma~\ref{prop:member} constrains the multipliers but says nothing about the candidate inputs, because the multipliers are themselves free unknowns.

\begin{lemma}[The multiplier scale is a gauge]\label{lem:magfree}
The residual $\br(\cdot)$ is positively homogeneous of degree one in $\lambda$, and the constraint
$\lambda\ge 0$ is preserved by $\lambda\mapsto c\lambda$ for $c>0$. Hence for any $z$ with
$\br(z)\neq\mathbf 0$ there is a unique $c>0$ whose rescaled multipliers satisfy
$\|\br\|_\dstar=\|\thv\|$ exactly. The magnitude bound therefore restricts no $\{\x_i\}$, and
\eqref{eq:member} is equivalent, modulo this one-dimensional gauge, to the cone membership
\begin{equation}\label{eq:cone}
\begin{aligned}
    \br(z) \in C(\thv) &:= \cone\big(\partial(\tfrac12\|\thv\|^2)\big) \\
    &= \{\,t\,\bk : \bk\in\partial(\tfrac12\|\thv\|^2),\, t\ge 0\,\}.
\end{aligned}
\end{equation}
\end{lemma}

Thus the magnitude of $\br$ is a scale convention, not a property of the data: everything the stationarity condition says about the training set is carried by the \emph{direction} of $\br$, i.e.\ by the cone $C(\thv)$, the central object of Section~\ref{sec:theory}. A reconstruction loss should therefore be invariant under $\lambda\mapsto c\lambda$; otherwise the attack can lower the loss merely by rescaling the multipliers (Lemma~\ref{prop:scale}). The second rescaling, $\thv\mapsto c\,\thv$, must be treated differently because the attacker cannot rescale the given $\thv$ (Lemma~\ref{claim:scale}, Appendix~\ref{app:moved}).

\paragraph{A layerwise Euler identity.} Blockwise homogeneity ties every residual block to one shared scalar, for any candidate and not only at solutions.

\begin{lemma}[Layerwise Euler identity]\label{lem:euler}
Under Assumption~\ref{ass:blockhom}, for every $z$ and every block $k$,
\begin{equation}\label{eq:euler}
\inner{\rr_{W_k}(z)}{W_k}=d_k\,\Gamma(z),\qquad \Gamma(z):=\sum_{i=1}^m\lambda_iy_if(\x_i;\thv).
\end{equation}
If moreover $\rr(z)\in\Cc(\thv)$, then $\Gamma(z)=\inner{\rr(z)}{\thv}/L>0$, hence
$\rr_{W_k}(z)\neq\mathbf0$ and $W_k\neq\mathbf0$ for every $k$; for a solution of \eqref{eq:member},
$\Gamma(z)=\Vert{}\thv\Vert{}^2/L$.
\end{lemma}

For the max-type norms of Muon and Adam, this identity implies that every block is active at a KKT direction (Corollary~\ref{cor:active}, Appendix~\ref{app:moved}).

\paragraph{From cone membership to a loss.} Membership in $C(\thv)$ is equivalent to tightness in the generalized Cauchy--Schwarz inequality \eqref{eq:aligndef}, so we penalize the relative Cauchy--Schwarz gap block by block, with block norms $\Vert{}\cdot\Vert{}_{(k)}$ and their duals,
\begin{equation}\label{eq:align}
\begin{split}
L_{\mathrm{align}} &= \sum_{k=1}^K\Big(1-\frac{\inner{\rr_{W_k}}{W_k}}{\|\rr_{W_k}\|_{(k),\dstar}\,\|W_k\|_{(k)}}\Big)^{2}, \\
&\quad \text{defined for } r_{W_k} \neq 0 \text{ for every } k.
\end{split}
\end{equation}

Each term lies in $[0,4]$ and vanishes iff $\rr_{W_k}$ is dual-aligned with $W_k$, i.e.\ $\rr_{W_k}\in C_k(\thv)\setminus\{\mathbf0\}$, where $C_k(\thv):=\cone\partial\Vert{}W_k\Vert{}_{(k)}$. The numerator and denominator are positively homogeneous of degree one in each argument, so the loss is invariant under the gauge of Lemma~\ref{lem:magfree} and requires no choice of singular vectors. By Lemma~\ref{lem:euler}, $\rr_{W_k}\neq\mathbf0$ in every block for any cone-feasible $z$, so this restriction discards no valid solution (for numerical stability we add a small $\epsilon$ to the denominator). Plugging in the appropriate $\Vert{}\cdot\Vert{}_{(k)}$ covers all three attacks with one formulation.

\begin{proposition}[Per-block alignment is exact]\label{prop:exact}
Let Assumption~\ref{ass:blockhom} hold, let $\Vert{}\cdot\Vert{}$ be one of the norms in \eqref{eq:norms} (Muon
with arbitrary $c_k>0$), and let \eqref{eq:member} have a solution at $\thv$. Then
\begin{align*}
\{z:\ \rr(z)\in\Cc(\thv)\} 
&= \{z:\ \rr_{W_k}(z)\in C_k(\thv)\setminus\{\mathbf0\}\ \ \forall k\} \\
&= \{z:\ L_{\mathrm{align}}(z)=0\}. 
\end{align*}
In particular, the zero set of $L_{\mathrm{align}}$ does not depend on the Muon weights $c_k$.
\end{proposition}

Hence no cross-block coupling term is needed (Remark~\ref{rem:couple}, Appendix~\ref{app:moved}).

To mount the attack we set $L_{\mathrm{stat}}:=L_{\mathrm{align}}$ in \eqref{eq:haim}, keeping $L_\lambda$ and $L_{\mathrm{prior}}$ unchanged. For $\ell_2$, Proposition~\ref{prop:exact} gives $L_{\mathrm{stat}}=0$ iff $\rr=c\,\thv$ for a shared $c>0$, so up to the gauge of Lemma~\ref{lem:magfree} the Euclidean case recovers the objective of \citet{haim2022}.

\subsection{The Muon (spectral) objective}
\label{sec:obj:muon}

Muon performs steepest descent under the spectral norm: for each block $W$,
$\|\cdot\|=\|\cdot\|_{\sms}=\sigma_1(\cdot)$, with dual $\|\cdot\|_{\nuc}$. Its subdifferential is a spectrahedron
rather than a point: if the top singular value has multiplicity $\rho$, with left and right singular
subspaces spanned by the columns of $U_1\in\R^{m\times\rho}$ and $V_1\in\R^{n\times\rho}$, then
\begin{equation}\label{eq:spectrahedron}
\partial\|W\|_{\sms}=\{\,U_1 M V_1\T:\ M\succeq0,\ \tr M=1\,\},
\end{equation}
which reduces to the rank-one $\pp\qq\T$ only when $\rho=1$. A natural first instinct is to keep the
template of \eqref{eq:haim} with a spectral target, minimizing the ``single-target'' loss
$\|\kk^\star-\rr_W\|_2^2$ for $\kk^\star=\nabla(\tfrac12\|W\|_{\sms}^2)=\sigma_1\,\pp\qq\T$. This breaks down for two
reasons: \emph{within} a block,
a repeated top singular value makes the target basis-dependent, and \emph{across} blocks, the relative
block scales it fixes need not match those realized at a KKT point (Lemma~\ref{prop:scale},
Appendix~\ref{app:moved:scale}). The Muon instance of
\eqref{eq:align}, in which $c_k$ cancels, reads
\begin{equation}\label{eq:muon-loss}
L_{\mathrm{align}}^{\mathrm{Muon}}=\sum_k\Big(1-\frac{\inner{\rr_{W_k}}{W_k}}{\|\rr_{W_k}\|_{\nuc}\,\|W_k\|_{\sms}}\Big)^{2}.
\end{equation}

Since every block attains the maximum at a KKT direction (Corollary~\ref{cor:active}), the extra
tightness condition of the max-type norm is vacuous (Lemma~\ref{lem:blockdecomp}) and
\eqref{eq:muon-loss} is exact for arbitrary per-block weights (Proposition~\ref{prop:exact}): no
balancing assumption is needed, and every layer carries information.

\subsection{The Adam (\texorpdfstring{$\ell_\infty$}{l\_infinity}) objective}
\label{sec:obj:adam}

Adam with a negligible stability constant maximizes the \emph{global} $\ell_\infty$ margin, so
$\|\thv\|=\|\thv\|_\infty$ with dual $\|\cdot\|_1$. With $A=\argmax_j|\theta_j|$ and $s=|A|$, H\"older
tightness holds iff $\rr_j=0$ for $j\notin A$ and $\rr_j\theta_j\ge0$ for $j\in A$, with $\rr\neq0$:
the candidate gradient combination must be supported on the largest-magnitude weights with matching
signs, and nothing else is asked of it. We use
the single global term 
\begin{equation}\label{eq:adam_stats_loss}
    (1-\inner{\rr}{\thv}/(\|\rr\|_1\|\thv\|_\infty))^2
\end{equation}
 (a per-tensor variant
is discussed in Appendix~\ref{app:moved:pertensor}). Why this target is substantially less weight-specific is the subject
of the next section.

\section{How much do the Optimizers Leak? A Specificity Analysis}
\label{sec:theory}

The attack seeks $z$ whose residual lies in $\Cc(\thv)$, and succeeds to the extent that this
\emph{forces} the candidates to be the training set. Counting the scalar equations that
$\rr\in C(\thv)$ imposes at a generic $\thv$ gives $\Theta(p)$ for all three geometries and cannot
explain the ordering (Appendix~\ref{app:count}); we instead ask how much of $\thv$ the
optimizer-dependent target $C(\thv)$ encodes.

\subsection{Weight-specificity and target-cone variability}
\label{sec:theory:spec}

Counting tells us how many equations there are, not how much they say about the weights. The
constraint references the weights through the residual map $z\mapsto\rr(z;\thv)$, fixed by the
architecture and $\thv$, and through the target $C(\thv)$, the only place where the optimizer enters.

\begin{definition}[Weight-specificity]\label{def:spec}
For a margin norm $\|\cdot\|$ and a stratum $\Sigma$, the \emph{weight-specificity}
$S_\Sigma(\|\cdot\|)$ is the dimension of the image of $\thv\mapsto C(\thv)$ restricted to the points of
$\Sigma$ at which all blocks are active, i.e.\ the number of independent continuous functionals of
$\thv$ that the stationarity target determines.
\end{definition}

Specificity is a property of the geometry and the stratum, not of a dataset: it counts the continuous
degrees of freedom through which the target cone varies with the weights, and is not by itself an
information-theoretic bound on input identifiability.

\begin{lemma}[Specificity of the three geometries]\label{lem:spec}
On the stratum with multiplicities $\rho_k$ and arg-max set $A$, at points where all blocks are active,
\begin{equation}\label{eq:spec}
S = \begin{cases}
p-1, & \ell_2,\\
\sum_k S_k(\rho_k), & \text{spectral},\\
0, & \ell_\infty,
\end{cases}
\end{equation}
since $C$ is the ray through $\thv$, determined by $\thv/\|\thv\|_2$ ($\ell_2$);
$C_k=\{U_{1,k}MV_{1,k}\T:M\succeq0\}$, determined by $U_{1,k}V_{1,k}\T$ (spectral); and
$C=\{\rr:\rr_j=0\text{ for }j\notin A,\ \rr_j\theta_j\ge0\text{ for }j\in A\}$ ($\ell_\infty$). Here $S_k(\rho):=\rho(m_k+n_k-\rho)-\rho(\rho+1)/2$ is the dimension of the manifold of rank-$\rho$
partial isometries in $\R^{m_k\times n_k}$; $S_k(1)=m_k+n_k-2$. In the $\ell_\infty$ case $C$ depends on
$\thv\neq\mathbf0$ only through the discrete data $(A,\sgn\theta|_A)$, for \emph{every} multiplicity
$s=|A|$.
\end{lemma}

\subsection{Consequences, and why Adam resists}
\label{sec:theory:cons}

For SGD the cone is the ray through $\thv$, so the constraint encodes the full direction
$\thv/\|\thv\|_2$: the candidate gradients must reproduce the entire weight vector up to one scalar.
This is the maximally specific geometry, and the regime in which \citet{haim2022} reconstruct. For
\textbf{Muon}, $C_k$ encodes the top singular structure $U_{1,k}V_{1,k}\T$ of \emph{every} block (all
are active at a KKT direction), i.e.\ $S_k(\rho_k)$ numbers per layer, at least $\Theta(m_k+n_k)$
(Appendix~\ref{app:specremark}). That is far less than $\ell_2$, but the constraint stays \emph{tight}: by
Lemma~\ref{lem:specfloor} many of each block's equations survive at any multiplicity in ordinary
matrix-block regimes, and it is attached to a weight-specific template that grows with width, so Muon
both constrains candidates sharply and carries a genuine data fingerprint. For \textbf{Adam}, with
$A = \argmax_l|\theta_l|$,
\begin{equation}\label{eq:linf-subdiff}
\partial\big(\tfrac12\|\thv\|_\infty^2\big)
= \|\thv\|_\infty\,\conv\{\sgn(\theta_j)\ee_j:\ j\in A\},
\end{equation}
an $(s-1)$-simplex whose \emph{interior directions are unconstrained}: the condition pins the support
and admissible signs of $\rr$ but never the relative magnitudes of $\thv$. Hence $S=0$
(cf.\ Remark~\ref{rem:linear}).

For networks the residual map carries $\thv$-dependence of its own, but the polyhedral geometry has a
second consequence that does extend: minimizing $\|\thv\|_\infty$ under margin constraints drives
$\thv$ onto a large face of the $\ell_\infty$ ball, where the constraint has few normal equations.
Under a first-order nondegeneracy condition (Assumption~\ref{ass:nondeg}), Theorem~\ref{thm:linf}
(Appendix~\ref{app:moved:linf}) shows that $\codim C(\thv)=p-s\le n_I$, where $n_I$ is the number of
on-margin examples. Adam's stationarity system thus has at most $n_I$ normal equations for $n_I(d+1)$
unknowns and, when $n_Id\ge2$, fails regular first-order identifiability; under additional
constant-rank and relative-interior conditions, the stationarity-consistent perturbations of the
on-margin candidates form a $C^1$ manifold of dimension at least $n_Id$. The result concerns the
idealized $\ell_\infty$ bias (Adam with a negligible stability constant) and this stationarity-only
attack family only, so it is not a differential-privacy-style guarantee \citep{abadi2016}.

\subsection{Two axes, and the vulnerability ordering}
\label{sec:theory:order}

Specificity concerns a \emph{converged} model, but the attack inverts a condition that holds only in the
limit, so a second axis matters: \emph{reachability}, how quickly an optimizer's parameter direction
approaches the KKT direction of its own margin problem. Muon is high on both axes: its specificity is
at least $\Theta(\text{width})$ in every Muon-trained layer, and per-block orthogonalization escapes the
small-initialization stall that traps gradient descent. SGD has the \emph{highest} specificity, $p-1$,
but the \emph{slowest} reachability (\citealp{haim2022} train ${\sim}10^6$ iterations): vulnerable in
principle, expensive in practice. Adam has fast reachability but zero continuous target
specificity and, at nondegenerate KKT points, too few equations for regular first-order identification
(Theorem~\ref{thm:linf}). Folding in cost, the \emph{practical} ordering is
\[
\boxed{\ \textbf{Muon}\ \succ\ \textbf{SGD}\ \succ\ \textbf{Adam}.\ }
\]
The Muon order is a prediction from reachability, not a theorem: that gap is about training
\emph{cost} and should narrow wherever SGD has budget to converge, while the Adam gap reflects weaker
continuous target dependence and regular underdetermination (summary in Table~\ref{tab:summary},
Appendix~\ref{app:proofs}).

\newcommand{\itersSGD}{10^{6}}       
\newcommand{\itersAdaptive}{10^{4}}  

\begin{figure*}[t]
\centering
\begin{subfigure}[b]{0.32\textwidth}
  \centering
  \includegraphics[width=\linewidth]{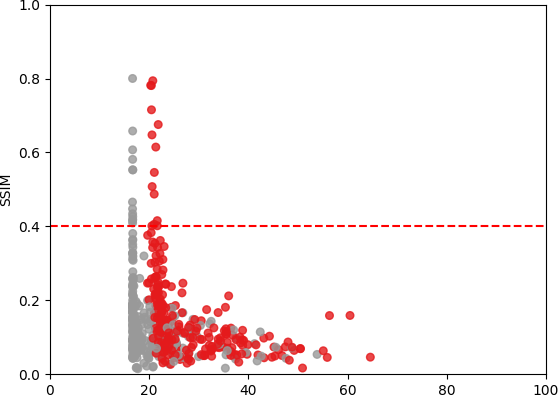}
  \caption{Muon, binary}
\end{subfigure}
\begin{subfigure}[b]{0.32\textwidth}
  \centering
  \includegraphics[width=\linewidth]{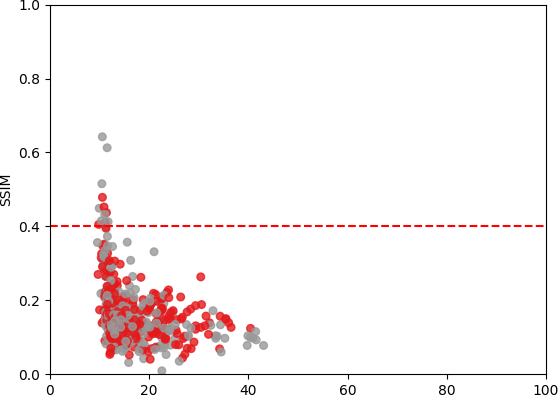}
  \caption{SGD, binary}
\end{subfigure}
\begin{subfigure}[b]{0.32\textwidth}
  \centering
  \includegraphics[width=\linewidth]{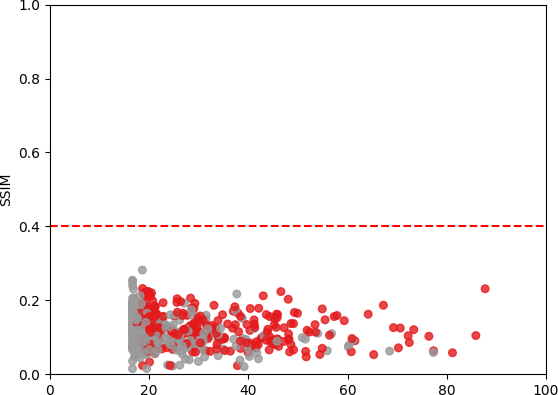}
  \caption{Adam, binary}
\end{subfigure}

\caption{Per-sample SSIM versus training margin \(\gamma_i\) on CIFAR-10 (\(n=500\), range prior), for binary models. Each point is a training sample, colored by class; the dashed line marks \(\mathrm{SSIM}=0.4\). Recognizable reconstructions occur near the minimal margin of their class. Muon produces most of them, and Adam essentially none.}
\label{fig:ssim-margin}
\end{figure*}

\section{Experiments}
\label{sec:exp}

We evaluate the attack of Section~\ref{sec:obj} on networks trained with SGD, Adam, and Muon and test
the predictions of Section~\ref{sec:theory}. Implementation details are in Appendix~\ref{app:details}
and full results in Appendix~\ref{app:extra}.

\paragraph{Setup.}\label{sec:exp:setup}
Unless stated otherwise, we use $n=500$ examples from binary MNIST \cite{lecun2010mnist} parity, binary CIFAR-10 \cite{krizhevsky2009learning}
vehicles versus animals, and 10-class CIFAR-10. Models are fully connected ReLU networks
$d\!-\!1000\!-\!1000\!-\!C$ ($C=1$ binary, $C=10$ multiclass), with only a first-layer bias,
absorbed into $W_1$, so Assumption~\ref{ass:blockhom} holds. SGD and Adam update all layers;
Muon applies five-step Newton--Schulz Muon \citep{jordan2024muon} to hidden-to-hidden
matrices and SGD to the input and output layers. We use the binary SGD models of
\citet{haim2022}. Models trained by Adam and Muon reach training loss $O(10^{-7})$ within
$\itersAdaptive$ iterations, whereas to reach the same order of training loss, SGD requires $\itersSGD$ iterations
(Table~\ref{tab:train}, Appendix~\ref{app:details}).

\paragraph{Attack and evaluation.}
The attack uses the Euclidean loss \eqref{eq:haim} for SGD, and the global $\ell_\infty$ alignment
loss (\S\ref{sec:obj:adam}) for Adam, and a smooth relaxation of \eqref{eq:muon-loss} for Muon. It optimizes $m=2n$ candidates with fixed balanced labels and their multipliers;
the multiclass extension is described in Appendix~\ref{app:details}. Unless stated otherwise,
no image prior is used. Reconstructions are matched one-to-one to training images by the
Hungarian algorithm on centered, normalized $\ell_2$ distance (more details in Appendix \ref{app:hung}). We evaluate SSIM@30 as our primary leakage metric. All-pairs averages can favor Adam because its candidates are generic images; @30 measures whether individual samples are recovered. We report SSIM, PSNR, and
$\ell_1$ and $\ell_2$ distances over all pairs and the best $30$ pairs (@30);
$\mathrm{SSIM}>0.4$ is considered visually recognizable.

\subsection{Main Results}
\label{sec:exp:main}

\paragraph{Muon-trained models are the most vulnerable.} Muon leads to the best reconstructions on every metric and task as visually illustrated for the binary classification setup in Figure~\ref{fig:optimizer_comparison}. Its SSIM@30 is $0.431$ on MNIST, $0.551$ on binary CIFAR-10 and $0.534$ on 10-class CIFAR-10, against $0.346$, $0.229$ and $0.383$ for SGD (the 10-class gap is discussed in Appendix~\ref{app:extra}). The gap is largest on binary
CIFAR-10, where Muon more than doubles the SGD score. Since no prior is used, these
reconstructions come from the stationarity condition alone, consistent with the fact that the spectral objective \eqref{eq:muon-loss} inverts the Muon condition and that the per-layer spectral template of Lemma~\ref{lem:spec} is exploitable, with two orders of magnitude fewer training iterations than SGD. This is confirmed by the sorted SSIM curves without priors across both CIFAR-10 and MNIST (Figure~\ref{fig:ssim_sorted_comparison}).

\paragraph{Adam-trained models are the most resistant.} Adam has the lowest SSIM@30 on every task without a prior, at most $0.249$, well below the $0.4$ recognizability level, consistent with the analysis of \S\ref{sec:theory:cons}
: Adam's $\ell_\infty$ target carries no
continuous information about the weight magnitudes, leaving the stationarity system too weakly
constrained to identify individual samples.

\paragraph{Ablation Studies: The effect of image priors.} \label{sec:exp:ablation}
 We add five regularizers to the attack objective: a range prior penalizing pixel values outside $[0,1]$, total variation (TV), an $\ell_1$ penalty favoring low-energy backgrounds, TV$+\ell_1$, and all three combined. As the only optimizer-independent component of the objective, the prior's marginal effect measures how much of a reconstruction does not come from the weights (Table~\ref{tab:priors}; full results in Tables~\ref{tab:full-mnist}-\ref{tab:full-multi}), consistent with recent findings on how priors can dominate checkpoint reconstruction fidelity \cite{refael2026no}. Qualitative comparisons showing reconstructions with and without priors are provided in Figure~\ref{fig:qual-mnist} for MNIST and Figure~\ref{fig:qual-cifar} for CIFAR-10.

\begin{table}[htb]
\centering
\caption{SSIM@30 ($\uparrow$) under different image priors ($n=500$). Best per task and prior in
\textbf{bold}, second best \underline{underlined}. Muon is best in all $18$ settings, and in all
$144$ entries of the full tables (Appendix~\ref{app:extra}).}
\label{tab:priors}
\small
\setlength{\tabcolsep}{4pt}
\resizebox{\linewidth}{!}{
\begin{tabular}{l ccc ccc ccc}
\toprule
 & \multicolumn{3}{c}{MNIST (binary)} & \multicolumn{3}{c}{CIFAR-10 (binary)} & \multicolumn{3}{c}{CIFAR-10 (10-class)}\\
\cmidrule(lr){2-4}\cmidrule(lr){5-7}\cmidrule(lr){8-10}
Prior & SGD & Adam & Muon & SGD & Adam & Muon & SGD & Adam & Muon\\
\midrule
None        & \underline{0.3463} & 0.2487 & \textbf{0.4310} & \underline{0.2287} & 0.1379 & \textbf{0.5508} & \underline{0.3827} & 0.2066 & \textbf{0.5339}\\
Range       & \underline{0.4057} & 0.2994 & \textbf{0.6371} & \underline{0.2336} & 0.1367 & \textbf{0.6130} & \underline{0.4515} & 0.1771 & \textbf{0.5315}\\
TV          & \underline{0.3768} & 0.2882 & \textbf{0.5808} & \underline{0.2574} & 0.2115 & \textbf{0.6132} & \underline{0.4332} & 0.3442 & \textbf{0.5481}\\
$\ell_1$    & \underline{0.3614} & 0.1752 & \textbf{0.5302} & \underline{0.2375} & 0.2112 & \textbf{0.5598} & \underline{0.4141} & 0.2739 & \textbf{0.5371}\\
TV$+\ell_1$ & \underline{0.3977} & 0.2311 & \textbf{0.6211} & \underline{0.2440} & 0.2369 & \textbf{0.6286} & \underline{0.4343} & 0.3050 & \textbf{0.5890}\\
All         & \underline{0.4511} & 0.2655 & \textbf{0.6599} & \underline{0.2536} & 0.2377 & \textbf{0.6189} & \underline{0.4343} & 0.2603 & \textbf{0.5795}\\
\bottomrule
\end{tabular}}
\end{table}

Priors do not change the leading optimizer: Muon is best under every prior (SSIM@30 between
$0.431$ and $0.660$), while Adam never exceeds $0.344$. On MNIST, combining all priors raises SSIM@30 by
$0.229$ for Muon and $0.105$ for SGD but only $0.017$ for Adam: priors sharpen reconstructions when
the stationarity condition already constrains them. On binary CIFAR-10, smoothness priors help Adam
most (from $0.138$ to $0.238$ with all priors). We
attribute this to smoothness priors making generic natural-image candidates more similar to the
training set, rather than to identification of specific samples, and to SGD not having reached its
limiting direction on this task (Table~\ref{tab:train}).

\paragraph{Effect of sample size and architecture.} To remove the training-budget asymmetry, we retrain
all models on binary CIFAR-10 to a training loss below $10^{-8}$, varying $n\in\{50,100,500,1000\}$ or
the hidden widths, and attack them with the range prior (Table~\ref{tab:abl}).

\begin{table}[htb]
\centering
\caption{Effect of sample size (top, $1000\!-\!1000$ hidden units) and architecture (bottom, $n=500$)
on binary CIFAR-10. All models are trained to a training loss below $10^{-8}$; the attack uses the
range prior. Best per configuration in \textbf{bold}.}
\label{tab:abl}
\small
\setlength{\tabcolsep}{3.5pt}
\resizebox{\linewidth}{!}{
\begin{tabular}{llcccccccc}
\toprule
Configuration & Optimizer & SSIM$\uparrow$ & SSIM@30$\uparrow$ & PSNR$\uparrow$ & PSNR@30$\uparrow$ & $\ell_1\downarrow$ & $\ell_1$@30$\downarrow$ & $\ell_2\downarrow$ & $\ell_2$@30$\downarrow$\\
\midrule
\multirow{3}{*}{$n=50$}
 & SGD      & 0.3024 & 0.4486 & 13.188 & 15.066 & 0.1586 & 0.1250 & 0.2191 & 0.1765\\
 & Adam     & 0.1452 & 0.1809 & 11.733 & 12.663 & 0.1989 & 0.1808 & 0.2590 & 0.2327\\
 & Muon & \textbf{0.4070} & \textbf{0.5848} & \textbf{14.409} & \textbf{17.390} & \textbf{0.1353} & \textbf{0.0954} & \textbf{0.1904} & \textbf{0.1351}\\
\midrule
\multirow{3}{*}{$n=100$}
 & SGD      & 0.2949 & 0.6232 & 13.499 & 18.765 & 0.1554 & 0.0828 & 0.2114 & 0.1153\\
 & Adam     & 0.0976 & 0.1488 & 11.214 & 12.570 & 0.2087 & 0.1775 & 0.2750 & 0.2352\\
 & Muon & \textbf{0.3501} & \textbf{0.6984} & \textbf{14.257} & \textbf{21.465} & \textbf{0.1404} & \textbf{0.0639} & \textbf{0.1937} & \textbf{0.0845}\\
\midrule
\multirow{3}{*}{$n=500$}
 & SGD      & 0.0994 & 0.2336 & 11.665 & 13.435 & 0.2027 & 0.1634 & 0.2611 & 0.2129\\
 & Adam     & 0.1299 & 0.1367 & 12.334 & 13.441 & 0.1863 & 0.1624 & 0.2417 & 0.2128\\
& Muon & \textbf{0.1921} & \textbf{0.613} & \textbf{12.8084} & \textbf{19.1901} & \textbf{0.175} & \textbf{0.0825} & \textbf{0.2289} & \textbf{0.1098}\\
\midrule
\multirow{3}{*}{$n=1000$}
 & SGD      & 0.0942 & 0.3001 & 11.537 & 14.168 & 0.2059 & 0.1476 & 0.2650 & 0.1957\\
 & Adam     & 0.0783 & 0.1202 & 11.150 & 11.956 & 0.2149 & 0.1968 & 0.2770 & 0.2525\\
 & Muon & \textbf{0.1442} & \textbf{0.4186} & \textbf{12.203} & \textbf{15.559} & \textbf{0.1895} & \textbf{0.1252} & \textbf{0.2454} & \textbf{0.1667}\\
\midrule\midrule
\multirow{3}{*}{$100\!-\!100$}
 & SGD      & 0.1132 & 0.1754 & 11.878 & 13.373 & 0.1975 & 0.1634 & 0.2547 & 0.2145\\
 & Adam     & 0.0569 & 0.0836 & 10.795 & 12.129 & 0.2242 & 0.1912 & 0.2886 & 0.2475\\
 & Muon & \textbf{0.1195} & \textbf{0.1826} & \textbf{11.921} & \textbf{14.001} & \textbf{0.1957} & \textbf{0.1598} & \textbf{0.2535} & \textbf{0.2039}\\
\midrule
\multirow{3}{*}{$1000\!-\!500\!-\!100$}
 & SGD      & \textbf{0.1654} & 0.4170 & 12.577 & 15.990 & \textbf{0.1808} & 0.1197 & 0.2351 & 0.1587\\
 & Adam     & 0.0927 & 0.1226 & 11.489 & 12.318 & 0.2065 & 0.1836 & 0.2664 & 0.2422\\
 & Muon & \textbf{0.1689} & \textbf{0.4234} & \textbf{12.832} & \textbf{16.182} & \textbf{0.1719} & \textbf{0.1170} & \textbf{0.2263} & \textbf{0.1552}\\
\midrule
\multirow{3}{*}{$1000\!-\!500\!-\!100\!-\!50$}
 & SGD      & 0.1617 & 0.3944 & 12.594 & 15.882 & 0.1804 & 0.1196 & 0.2346 & 0.1607\\
 & Adam     & 0.1181 & 0.1546 & 11.944 & 12.736 & 0.1950 & 0.1741 & 0.2528 & 0.2308\\
 & Muon & \textbf{0.1896} & \textbf{0.4236} & \textbf{13.5685} & \textbf{17.962} & \textbf{0.1622} & \textbf{0.1085} & \textbf{0.2192} & \textbf{0.1396}\\
\bottomrule
\end{tabular}}
\end{table}

Across sample sizes, Muon is best on every metric and Adam has the lowest SSIM@30. Once SGD is
trained to convergence, however, it becomes competitive at small $n$: at $n=100$ its SSIM@30 is
$0.623$ against $0.698$ for Muon, compared with $0.234$ against $0.613$ at $n=500$. The Muon gap
is therefore partly a gap in training cost, as predicted by the reachability argument of
\S\ref{sec:theory:order}.

Adam is worst on every metric for every
architecture, with SSIM@30 below half that of the better optimizer. The Adam separation is thus robust
and the Muon comparison regime-dependent, as the theory suggests: Adam's weakness follows from the
structure of the $\ell_\infty$ target cone, whereas the comparison between the $\ell_2$ and spectral
geometries depends on the residual map and the training budget.

\paragraph{Which Samples Are Reconstructed?}
\label{sec:exp:analysis} By complementary slackness, only samples on the margin carry positive multipliers and enter the
stationarity condition. Figure~\ref{fig:ssim-margin} tests this by plotting, for each training sample,
the SSIM of its matched reconstruction against its margin $\gamma_i$. In both the binary and multiclass
settings (Figure~\ref{fig:ssim-margin-multi}), recognizable reconstructions occur, for every optimizer, only for samples at or near the
minimal margin, and samples well inside the margin are never recovered. Muon produces by far the
most samples above the threshold, many concentrated at the minimal margin (a narrow vertical band);
SGD produces a handful and Adam essentially none. Being on the margin is necessary for recovery under every geometry, but it is sufficient only when the stationarity condition carries
weight-specific information, which the $\ell_\infty$ condition does not.

\section{Discussion}
Our results identify optimizer implicit bias as an overlooked axis of training-data privacy. Even with identical architectures, data, and interpolation behavior, optimizers leave distinct geometric fingerprints: Muon’s spectral bias tightly constrains candidate inputs and yields the strongest reconstructions, SGD leaks less, and Adam is most resistant. This theoretically predicted ordering—driven by the specificity and reachability of stationarity constraints—shows that optimizer choice can serve as a practical privacy control, not merely an optimization decision. Although our checkpoint-only, stationarity-based analysis provides no formal privacy guarantee, it demonstrates that evaluating model privacy independently of the training algorithm can overlook a consequential source of leakage.

Finally, while our study focuses on standard implementations of SGD, Adam, and Muon, modern training pipelines increasingly adopt memory-efficient, low-rank, and orthogonalized optimizers. Extending geometric implicit-bias analyses to methods such as AdaRankGrad \cite{refael2025adarankgrad}, SUMO \cite{refael2025sumo}, \cite{refael2026lorenza}, and \cite{svirsky2026train} presents a promising direction to understand how adaptive rank reduction and subspace orthogonalization govern training-data leakage.

\subsection*{AI use statement}

In this work, we used generative AI tools for text polishing and obtaining suggestions on document organization and section structure.
We have not used generative AI tools for research idea generation, code development, mathematical derivations, or experimental data generation/analysis, and all other disclosure categories are not applicable to this work.
We have reviewed all AI-assisted work.
Specifically, all AI-suggested structural edits and polished text were thoroughly examined, manually refined, and verified for technical accuracy and consistency by the authors.
We take responsibility for the final content of this work, including text, claims, or artifacts produced with the aid of generative AI.

\bibliography{referance.bib}

\clearpage
\appendix
\thispagestyle{empty}

\onecolumn
\aistatstitle{Supplementary Materials}

\section{Additional Analysis of Reconstruction Objectives and Privacy Leakage}

\label{app:moved}

\subsection{Scale covariance in the trained weights (from \S\ref{sec:obj:gauge})}
Besides the multiplier gauge of Lemma~\ref{lem:magfree}, there is another rescaling invariance, on the
weights, that should be treated differently.

\begin{lemma}[Scale covariance in $\thv$]\label{claim:scale}
If $z=(\{\x_i\},\{\lambda_i\})$ satisfies \eqref{eq:member} at $\thv$, then for every $c>0$ the pair
$(\{\x_i\},\{c^{\,2-L}\lambda_i\})$ satisfies it at $c\,\thv$.
\end{lemma}

In line with \citet{haim2022}, Lemma~\ref{claim:scale} justifies disregarding the norm of $\thv$ in 
the homogeneous stationarity equation: every $\thv$ on the ray remains a valid stationary point up 
to a rescaling of the multipliers. Importantly, this invariance does not preserve primal feasibility 
or complementary slackness under the rigid margin constraint ``$\ge 1$'' in \eqref{eq:maximal_margin}.

\subsection{Every block is active (from \S\ref{sec:obj:gauge})}

For the max-type norms governing Muon and Adam, the Euler identity (Lemma~\ref{lem:euler}) enforces a
direct structural constraint.

\begin{corollary}[Every block is active at a KKT direction]\label{cor:active}
Let Assumption~\ref{ass:blockhom} hold and let \eqref{eq:member} have a solution at $\thv$.
\emph{(a)} For the Muon norm of \eqref{eq:norms}, every block attains the maximum:
$\Vert{}W_k\Vert{}_{\sms}/c_k=\Vert{}\thv\Vert{}$ for all $k$. \emph{(b)} For the Adam norm, every block contains a globally
maximal coordinate: $A\cap W_k\neq\emptyset$ and $\Vert{}W_k\Vert{}_\infty=\Vert{}\thv\Vert{}_\infty$ for all $k$, where
$A=\argmax_j\vert{}\theta_j\vert{}$.
\end{corollary}

\subsection{Why per-block alignment loses nothing (from \S\ref{sec:obj:gauge})}

At first glance, optimizing a per-block loss appears to relax the global condition. Under an $\ell_2$ norm, for example, it permits $\rr_{W_k}=t_kW_k$ using a distinct scalar $t_k>0$ for each block. For max-type norms, it seemingly overlooks which specific blocks actually attain the maximum. However, the Euler identity proves that at the exact weights an attacker encounters in a real scenario, this per-block approach remains completely exact (Proposition~\ref{prop:exact}).

\begin{remark}[No coupling term is needed]\label{rem:couple}
One might add $L_{\mathrm{couple}}=\operatorname{Var}_k\big[\log(\Vert{}\rr_{W_k}\Vert{}_{(k),\dstar}/\Vert{}W_k\Vert{}_{(k)})\big]$
to restore the relative block scales that a per-block loss seems to drop. On the zero set of
$L_{\mathrm{align}}$, \eqref{eq:euler} gives
$\Vert{}\rr_{W_k}\Vert{}_{(k),\dstar}\Vert{}W_k\Vert{}_{(k)}=\inner{\rr_{W_k}}{W_k}=d_k\Gamma(z)$, so
$L_{\mathrm{couple}}=\operatorname{Var}_k\big[\log(d_k/\Vert{}W_k\Vert{}_{(k)}^2)\big]$ is a function of $\thv$
alone. It therefore cannot discriminate between candidates satisfying $L_{\mathrm{align}}=0$, and we do
not use it.
\end{remark}

\subsection{Equation counting at a generic weight vector does not separate the geometries}
\label{app:count}\label{sec:theory:count}

The constraint $\rr\in C(\thv)$ imposes $\codim C(\thv)$ scalar equalities, plus sign constraints. By
Lemma~\ref{lem:subdiff} and Corollary~\ref{cor:active}, at a KKT direction on the stratum of $\thv$,
\begin{equation}\label{eq:codim}
\codim C(\thv)=
\begin{cases}
p-1, & \ell_2\ (C=\text{ray through }\thv),\\
\sum_k \big(m_kn_k-\rho_k(\rho_k+1)/2\big), & \text{spectral}\ (C=\textstyle\prod_k\{U_{1,k}MV_{1,k}\T:M\succeq0\}),\\
p-s, & \ell_\infty\ (C=\text{simplicial cone on }A,\ s=|A|).
\end{cases}
\end{equation}
At a generic $\thv$ ($\rho_k=1$, $s=1$) these are $p-1$, $p-K$ and $p-1$, all $\Theta(p)$ in the
usual nontrivial matrix-block regime, so counting at a generic weight vector explains no ordering. But
KKT points are not generic, and the geometries respond very differently to degeneracy. For the
spectral norm, higher multiplicity costs few equations:

\begin{lemma}[The spectral constraint cannot collapse]\label{lem:specfloor}
For every $\thv$ and all multiplicities, with $r_k=\min(m_k,n_k)$,
\[
\textstyle\sum_k\big(m_kn_k-\rho_k(\rho_k+1)/2\big)\ \ge\ \tfrac12\big(p-\sum_kr_k\big).
\]
\end{lemma}

For $\ell_\infty$, in contrast, $p-s$ can be as small as $0$, and \S\ref{sec:theory:cons} shows that at
the KKT points Adam reaches it is at most the number of on-margin examples. The relevant comparison is
with the unknowns. Each on-margin candidate contributes $d+1$ of them (input and multiplier), and the
true data can satisfy the regular first-order identifiability criterion only if
$\codim C(\thv)\ge n_I(d+1)-1$, where $n_I$ is the number of on-margin examples
(Lemma~\ref{lem:Bnec}). This is necessary, not sufficient: off-margin candidates contribute nothing
($\lambda_i=0$), the prior and $\lambda\ge0$ add restrictions, and recovering a \emph{subset} of the
points needs no unique global solution. The $\ell_2$ and spectral geometries meet it with room to spare
in the over-parameterized regime $p\gg nd$, and it is exactly what fails for $\ell_\infty$.

\subsection{Target-cone sufficiency for linear-in-parameter models (from \S\ref{sec:theory:spec})}

When the residual map does not depend on $\thv$, the checkpoint affects
the stationarity-feasible set only through the target cone. This occurs for models linear in their
parameters, $f(\x;\thv)=\inner{\thv}{\phi(\x)}$ for a fixed feature map $\phi$ (random-feature, kernel
and linearized-network models), where $\rr(z)=\sum_i\lambda_iy_i\phi(\x_i)$.

\begin{lemma}[Target-cone sufficiency for linear-in-parameter models]\label{lem:ceiling}
Let $f(\x;\thv)=\inner{\thv}{\phi(\x)}$. If $C(\thv)=C(\thv')$, then the stationarity-feasible sets
\[
\{z:\rr(z)\in\Cc(\thv)\}
\quad\text{and}\quad
\{z:\rr(z)\in\Cc(\thv')\}
\]
coincide, as do the zero sets of the corresponding global alignment losses. Hence, within the
stationarity-only formulation, the checkpoint affects feasibility only through $C(\thv)$. On each
stratum these cones vary in an $S$-dimensional continuous family. This invariance does not alone imply
non-identifiability of the inputs, which also depends on the residual map $\phi$ and on the codimension
of the cone.
\end{lemma}

For such models $C(\thv)$ is a sufficient statistic of the weights for this stationarity-only
formulation. For networks, $\rr(z;\thv)=\sum_i\lambda_iy_i\nabla_{\thv}f(\x_i;\thv)$ depends on $\thv$
as well, so two networks with equal cones but different weights have different residual maps and
$C(\thv)$ is no longer sufficient. Specificity then measures only the part of the constraint that the
optimizer controls; the network-level regular first-order result is Theorem~\ref{thm:linf}.
Attackers with extra information (checkpoints, query access, a strong generative prior) are outside the
scope of either.

\subsection{Stratum invariance under \texorpdfstring{$\ell_\infty$}{l\_infinity} (from \S\ref{sec:theory:cons})}

\begin{proposition}[Stratum invariance under $\ell_\infty$]\label{prop:nonid}
Let $f(\x;\thv)=\inner{\thv}{\phi(\x)}$. The stationarity-feasible set for the Adam geometry depends
on $\thv$ only through $(A,\sgn\thv|_A)$. Consequently, checkpoints sharing this arg-max/sign pattern
induce exactly the same stationarity equations and the same feasible candidate datasets. Thus the
target contributes no continuous dependence on the weight magnitudes within a fixed stratum. This
invariance does not by itself imply non-identifiability of the inputs, since the equations
$\rr(z)\in C(\thv)$ may still constrain or identify $z$ through the feature map $\phi$.
\end{proposition}

For these models Adam's target cone transmits an $O(p)$-bit combinatorial arg-max/sign pattern but no
continuous information about the weight magnitudes within a fixed stratum. This makes the target less
weight-specific, although the resulting stationarity equations may still constrain the inputs through
the feature map.

\subsection{Discussion of Theorem~\ref{thm:linf} (from \S\ref{sec:theory:cons})}
\label{app:moved:linfdisc}

On Assumption~\ref{ass:nondeg}: The span condition says that $\thv$ is first-order isolated given its active set: no direction moves
$\thv$ while preserving all tied maxima and all active margins to first order. For
linear-in-parameter models it holds whenever the $\ell_\infty$ max-margin linear program has a unique
solution, since a unique LP solution is a vertex (Remark~\ref{rem:lp}). It is checkable on a trained
model through its consequence $s\ge p-n_I$.

Thus, under Assumption~\ref{ass:nondeg}, Adam's stationarity system has at most $n_I$ normal equations
for $n_I(d+1)$ unknowns and fails regular first-order identifiability. Under the additional
constant-rank and relative-interior hypotheses of part (iii), this first-order deficiency becomes an
actual positive-dimensional family of stationarity-consistent candidates, whose input projection has
dimension at least $n_I(d-1)$. Without those additional hypotheses, the rank count alone does not
exclude an isolated solution. These conclusions concern the stationarity-only attack, not the full KKT
system or other reconstruction attacks. The $\ell_2$ and spectral constraints have many more normal
equations in ordinary over-parameterized matrix-block regimes (Lemma~\ref{lem:specfloor}). Three
caveats remain. The results concern the idealized $\ell_\infty$ bias, i.e.\ Adam with a negligible
stability constant. The network statement rests on Assumption~\ref{ass:nondeg}, a property of the
trained model, and actual local ambiguity additionally rests on the hypotheses of
Theorem~\ref{thm:linf} (iii). Finally, they cover \emph{this} attack family only, so they are not a
differential-privacy-style guarantee \citep{abadi2016}.

Specificity and Theorem~\ref{thm:linf} explain why Adam is less exposed to this stationarity-only
attack, but they do not show that $\ell_2$ or spectral geometry \emph{succeeds}. Regular first-order
well-posedness in the attacker's variables is formalized in Appendix~\ref{app:local} as a rank condition
on the Jacobian of the constrained residual at $z^\star$. Of the resulting Conjecture~\ref{conj:jac},
the Adam failure of regular first-order identifiability follows from Theorem~\ref{thm:linf}
(Proposition~\ref{prop:Bneg}); for the positive half we prove only the necessary condition
$\codim C(\thv)\ge n_I(d+1)-1$, which the $\ell_2$ and spectral geometries satisfy in the stated
over-parameterized regime.

\subsection{Set-valued target and collapse (from \S\ref{sec:obj:muon})}
\label{app:moved:scale}

\emph{Degeneracy within blocks}: spectral-margin maximization drives blocks toward repeated top
singular values, and then $\pp,\qq$ are defined only up to rotation within the corresponding top
singular subspaces $U_1,V_1$. Thus the single-target construction is basis-dependent and is not
selected by the KKT condition.

\emph{Degeneracy across blocks}: by Corollary~\ref{cor:active} every block attains the maximum at a KKT direction, so
$\partial(\tfrac12\|\thv\|^2)$ is set-valued even when every $\rho_k=1$. It contains a $(K-1)$-simplex of relative
block weights, and the element that solutions actually realize is singled out only through the Euler
identity, with relative block scales that depend on the weights $c_k$ and the degrees $d_k$.

\begin{lemma}[Set-valued target and collapse]\label{prop:scale}
Let Assumption~\ref{ass:blockhom} hold and let \eqref{eq:member} have a solution at $\thv$ for the Muon
norm with weights $c_k$. Then:
\begin{enumerate}[label=(\alph*),leftmargin=*,nosep]
\item Each term of \eqref{eq:muon-loss} is invariant under $\rr_{W_k}\mapsto a_k\rr_{W_k}$ and
$W_k\mapsto b_kW_k$ for all $a_k,b_k>0$;
\item $\partial(\tfrac12\|\thv\|^2)=\big\{(\|\thv\|\,\mu_kc_k^{-1}G_k)_k:\ \mu\in\Delta_K,\ G_k\in\partial\|W_k\|_{\sms}\big\}$,
a convex set of dimension $(K-1)+\sum_k\big(\rho_k(\rho_k+1)/2-1\big)$, where $\Delta_K$ is the
probability simplex. Every solution of \eqref{eq:member} has
$\rr_{W_k}=\big(d_k\sigma_1(W_k)/(Lc_k^2)\big)G_k$ with $G_k\in\partial\|W_k\|_{\sms}$. Hence if all
$\rho_k=1$, solutions realize a single point $\kk^\circ$, which is a positive multiple of the naive
target $\kk^\star$ iff $d_k/c_k^2$ is constant in $k$; if some $\rho_k>1$, the KKT condition selects no
unique point target. Although $U_{1,k}V_{1,k}\T/\rho_k$ is a canonical basis-independent choice,
imposing it would add a restriction not implied by stationarity;
\item for any fixed target $\kk$ and direction $\sv$, on the ray $\rr=t\sv$, $t\ge0$, the fixed-target
loss $\|\kk-t\sv\|^2$ is minimized at $t^\star=(\inner{\kk}{\sv})_+/\|\sv\|^2$. So if
$\inner{\kk}{\sv}\le0$ the minimizer is $\rr=\mathbf0$, and otherwise the minimum is
$\|\kk\|^2\sin^2\angle(\kk,\sv)$, zero iff $\sv$ is a positive multiple of $\kk$. Each term of
\eqref{eq:muon-loss} is constant along the ray.
\end{enumerate}
\end{lemma}

Part (a) shows that the alignment loss is blind to both per-block scales, so by
Proposition~\ref{prop:exact} its zero set is the exact solution set, up to the gauge, for any $c_k$,
which the attacker need not know. Part (b) identifies what the naive loss gets wrong. Within a block
with $\rho_k>1$ it must select one element of a spectrahedron, even though stationarity does not specify
which element the true residual realizes. Across blocks its relative scales are correct only when
$d_k/c_k^2$ is constant; otherwise $L_{\mathrm{single}}=0$ has no solution at all, even for the true
data. Part (c) is a landscape property shared by every fixed-target loss, including the Euclidean
\eqref{eq:haim}: when the residual is negatively correlated with the target, the cheapest descent is to
send the multipliers to zero, switching the candidates off rather than aligning them with data. The
alignment loss is constant along multiplier rays and has no such incentive. Being restricted to one
ray, (c) is a mechanism rather than a proof of failure on the full domain.

\subsection{Remarks on Lemma~\ref{lem:spec} (from \S\ref{sec:theory:spec})}
\label{app:specremark}

Since $S_k$ is concave in $\rho$, $S_k(\rho_k)\ge\min\{S_k(1),S_k(r_k)\}$, and $S_k(r_k)\ge r_k(r_k-1)/2$.
In particular $S_k(\rho)\ge m_k+n_k-2$ for every $\rho$ whenever $r_k\ge4$: higher multiplicity never
erases Muon's template. Specificity thus separates the continuous target-cone variability of the
geometries where generic counting could not. It is also consistent with \S\ref{sec:obj}: the magnitudes
$\sigma_1(W_k)$ are absent from \eqref{eq:spec} because the cone is scale-free, the same fact that makes
the block scales of a fixed target a nuisance (Lemma~\ref{prop:scale}(b)). A loss cannot recover
continuous target information the constraint does not contain, although the fixed cone can still
constrain inputs through the residual map.

\begin{remark}[Linear-in-parameter models]\label{rem:linear}
For $f(\x;\thv)=\inner{\thv}{\phi(\x)}$ the residual map does not depend on $\thv$, so the checkpoint
affects the stationarity-feasible set only through $C(\thv)$ (Lemma~\ref{lem:ceiling}); under the Adam
geometry this set depends on $\thv$ only through $(A,\sgn\thv|_A)$ (Proposition~\ref{prop:nonid}).
Neither statement alone implies non-identifiability of the inputs (Appendix~\ref{app:moved}).
\end{remark}

\subsection{Collapse of the \texorpdfstring{$\ell_\infty$}{l\_infinity} constraint (from \S\ref{sec:theory:cons})}
\label{app:moved:linf}

For networks the residual map carries $\thv$-dependence of its own. The polyhedral
geometry, however, has a second consequence that does extend. Minimizing $\|\thv\|_\infty$ under margin
constraints drives $\thv$ onto a large face of the $\ell_\infty$ ball (for $a\T\thv\ge1$ with all
$a_j\neq0$, the solution $\sgn(a)/\|a\|_1$ has \emph{every} coordinate maximal), and on such a face
the constraint has few normal equations. We make this precise under a first-order nondegeneracy
condition. Let $z^\star$ be the true solution, $I=\{i:\lambda_i^\star>0\}$ and $n_I=|I|$.

\begin{assumption}[Nondegenerate $\ell_\infty$ KKT point]\label{ass:nondeg}
Strict complementarity holds (the examples with $\lambda_i^\star>0$ are exactly those on the margin),
and the vectors $\{\ee_j:\ j\in A\}\cup\{\nabla_{\thv}f(\x_i^\star;\thv):\ i\in I\}$ span $\R^p$.
\end{assumption}

\begin{theorem}[Collapse of the $\ell_\infty$ constraint]\label{thm:linf}
Under Assumption~\ref{ass:nondeg}, with $\rr$ continuously differentiable in $z$ near $z^\star$:
\begin{enumerate}[label=(\roman*),leftmargin=*,nosep]
\item $s\ge p-n_I$, hence $\codim C(\thv)=p-s\le n_I$;
\item let $z_I=(\{\x_i\},\{\lambda_i\})_{i\in I}\in\R^{n_I(d+1)}$, the remaining candidates being held at
$z^\star$, and $\Phi(z_I):=P_N\rr(z)$ with $P_N$ the projection onto the coordinates outside $A$. Then
$\rank D\Phi(z_I^\star)\le n_I$; if $n_Id\ge2$, this is less than $n_I(d+1)-1$, so $z^\star$ is not
regularly first-order identifiable (Definition~\ref{def:jac});
\item if moreover $D\Phi$ has constant rank near $z_I^\star$ and $\rr_j(z^\star)\neq0$ for $j\in A$, the
set of $z_I$ near $z_I^\star$ with $\lambda_i>0$ and $\rr(z)\in\Cc(\thv)$ is a $C^1$ manifold through
$z_I^\star$ of dimension at least $n_Id$. Arbitrarily close to $z_I^\star$, its projection onto the
input coordinates contains $C^1$ manifolds of dimension at least $n_I(d-1)$.
\end{enumerate}
\end{theorem}

See Appendix~\ref{app:moved:linfdisc} for a discussion of the assumption and of the result.

\subsection{Per-tensor form of the Adam objective (from \S\ref{sec:obj:adam})}
\label{app:moved:pertensor}

Besides the single global term, we also evaluate the \emph{per-tensor} form \eqref{eq:align} with
$\|\cdot\|_{(k)}$ the $\ell_\infty$ norm on block $k$.
For general $\thv$ the per-tensor form is a strict relaxation, since it permits a nonzero residual in
tensors containing no globally maximal coordinate. At a KKT direction every tensor contains one
(Corollary~\ref{cor:active}(b)), and the two forms have the same zero set
(Proposition~\ref{prop:exact}).

\section{Proofs}
\label{app:proofs}

Throughout, $\|\cdot\|$ is the optimizer's implicit-bias (margin) norm, $\|\cdot\|_\dstar$ its dual,
and $h(\thv)=\tfrac12\|\thv\|^2$. We collect two elementary lemmas (\S\ref{app:lemmas}), then prove the
membership test (\S\ref{app:member}), the scale statements and the Euler identity (\S\ref{app:scale}),
exactness of per-block alignment (\S\ref{app:exact}), the set-valued-target lemma
(\S\ref{app:coupling}), the block decomposition (\S\ref{app:blockdecomp}), and the specificity and
collapse results (\S\ref{app:proofs-count}).

\begin{table}[htb]
\centering
\caption{Summary. The cone $C(\thv)$ is the optimizer-dependent part of the constraint and, for
linear-in-parameter models, is the part through which the checkpoint affects stationarity feasibility
(Lemma~\ref{lem:ceiling}). Codimensions are at a KKT direction on the stratum of $\thv$; at a generic
$\thv$ all three are $\Theta(p)$ in the usual nontrivial matrix-block regime and separate nothing.
$S_k(\rho)=\rho(m_k+n_k-\rho)-\rho(\rho+1)/2$, $r_k=\min(m_k,n_k)$.}
\label{tab:summary}
\small
\begin{tabular}{lllll}
\toprule
optimizer & cone $C(\thv)$ & codim at KKT stratum & specificity $S$ & reachability\\
\midrule
SGD  & ray through $\thv$ & $p-1$ & $p-1$ & slow\\
Muon & $\prod_k\{U_{1,k}MV_{1,k}\T:M\succeq0\}$ & $\ge\tfrac12(p-\sum_kr_k)$ & $\sum_kS_k(\rho_k)$ & fast\\
Adam & simplicial cone on $A$, signs of $\thv|_A$ & $p-s\le n_I$ (Thm.~\ref{thm:linf}) & $0$ (combinatorial only) & fast\\
\bottomrule
\end{tabular}
\end{table}

\subsection{Two helper lemmas}
\label{app:lemmas}

\begin{lemma}[Subdifferential of a squared norm]\label{lem:half}
For any norm $\|\cdot\|$ with dual $\|\cdot\|_\dstar$ and any $\thv$,
\[
\partial h(\thv)=\|\thv\|\,\partial\|\thv\|,\qquad
\partial\|\thv\|=\{\uu:\ \|\uu\|_\dstar\le1,\ \inner{\uu}{\thv}=\|\thv\|\}.
\]
Moreover $\partial\|\cdot\|$ is $0$-homogeneous, so $\partial\|c\thv\|=\partial\|\thv\|$ and
$\partial h(c\thv)=c\,\partial h(\thv)$ for every $c>0$.
\end{lemma}
\begin{proof}
Write $h=g\circ N$ with $N=\|\cdot\|$ and $g(t)=\tfrac12 t^2$. The outer function $g$ is convex,
differentiable and nondecreasing on $[0,\infty)\supseteq\operatorname{range}(N)$, so the convex chain
rule gives $\partial h(\thv)=g'(N(\thv))\,\partial N(\thv)=N(\thv)\,\partial N(\thv)$. The displayed
form of $\partial N$ is standard, from $N(\thv)=\sup_{\|\uu\|_\dstar\le1}\inner{\uu}{\thv}$. For
$c>0$, $\uu\in\partial\|c\thv\|$ iff $\|\uu\|_\dstar\le1$ and
$\inner{\uu}{c\thv}=\|c\thv\|=c\|\thv\|$, i.e.\ iff $\uu\in\partial\|\thv\|$. Hence
$\partial h(c\thv)=c\|\thv\|\,\partial\|\thv\|=c\,\partial h(\thv)$.
\end{proof}

\begin{lemma}[Gradient homogeneity]\label{lem:gradhom}
If $f(\x;\cdot)$ is $L$-homogeneous and differentiable in $\thv$, then
$\nabla_{\thv} f(\x;\alpha\thv)=\alpha^{\,L-1}\nabla_{\thv} f(\x;\thv)$ for all $\alpha>0$.
\end{lemma}
\begin{proof}
Differentiate $f(\x;\alpha\thv)=\alpha^{L}f(\x;\thv)$ with respect to $\thv$: the left-hand side is
$\alpha\,\nabla_{\thv} f(\x;\alpha\thv)$ and the right-hand side $\alpha^{L}\nabla_{\thv} f(\x;\thv)$;
divide by $\alpha>0$. For ReLU networks the identity holds wherever the gradient exists, and everywhere
for the smooth surrogate used in the attack.
\end{proof}

\subsection{Membership test (Lemma~\ref{prop:member})}
\label{app:member}
\begin{proof}
($\Leftarrow$) The two hypotheses and the generalized Cauchy--Schwarz inequality
$\inner{\rr}{\thv}\le\|\rr\|_\dstar\|\thv\|$ chain into
$\|\thv\|^2\le\inner{\rr}{\thv}\le\|\rr\|_\dstar\|\thv\|\le\|\thv\|^2$, which forces all three to be
equalities. Thus $\|\rr\|_\dstar=\|\thv\|$ and $\inner{\rr}{\thv}=\|\thv\|^2$, so
$\vv:=\rr/\|\thv\|$ satisfies $\|\vv\|_\dstar=1$ and $\inner{\vv}{\thv}=\|\thv\|$, that is
$\vv\in\partial\|\thv\|$ and $\rr=\|\thv\|\vv\in\|\thv\|\,\partial\|\thv\|$ by
Lemma~\ref{lem:half}. ($\Rightarrow$) is immediate from the same two defining properties of
$\partial\|\thv\|$.
\end{proof}

\subsection{Scale statements and the Euler identity}
\label{app:scale}

\begin{proof}[Proof of Lemma~\ref{lem:magfree}]
The residual $\rr(z)=\sum_i\lambda_iy_i\nabla_{\thv}f(\x_i;\thv)$ is linear in $\lambda$, so replacing
$\lambda\mapsto c\lambda$ with $c>0$ gives $\rr\mapsto c\,\rr$ and preserves $\lambda\ge0$; if
$\rr\neq\mathbf0$, the choice $c:=\|\thv\|/\|\rr\|_\dstar$ achieves
$\|c\,\rr\|_\dstar=\|\thv\|$, and it is the only one. For the equivalence, fix the inputs.
($\Rightarrow$) If $\rr\in\partial h(\thv)$ then $\rr\in C(\thv)$, and $\rr\neq\mathbf0$ because
$\inner{\rr}{\thv}=\|\thv\|^2>0$ by Lemma~\ref{prop:member}. ($\Leftarrow$) If
$\rr\in\Cc(\thv)$ then $\rr=t\kk$ with $t>0$ and $\kk\in\partial h(\thv)$; replacing $\lambda$ by
$\lambda/t\ge0$ turns the residual into $\kk$, a solution of \eqref{eq:member}. No homogeneity of $f$
is used.
\end{proof}

\begin{proof}[Proof of Lemma~\ref{claim:scale}]
Fix $c>0$ and set $\lambda_i'=c^{\,2-L}\lambda_i\ge0$. By Lemma~\ref{lem:gradhom} with $\alpha=c$,
\[
\rr'=\sum_i\lambda_i'\,y_i\,\nabla_{\thv} f(\x_i;c\thv)
   =\sum_i c^{\,2-L}\lambda_i\,y_i\,c^{\,L-1}\nabla_{\thv} f(\x_i;\thv)=c\,\rr .
\]
By hypothesis $\rr\in\partial h(\thv)$, and by Lemma~\ref{lem:half}
$\partial h(c\thv)=c\,\partial h(\thv)$; hence $\rr'\in\partial h(c\thv)$, which together with
$\lambda_i'\ge0$ is \eqref{eq:member} at $c\thv$. This proves covariance of stationarity; primal
feasibility and complementary slackness for the fixed unit margin are not asserted.
\end{proof}

\begin{proof}[Proof of Lemma~\ref{lem:euler}]
Fix $\x$ and $k$. Differentiating
$\alpha\mapsto f(\x;W_1,\dots,\alpha W_k,\dots,W_K)=\alpha^{d_k}f(\x;\thv)$ at $\alpha=1$ gives
Euler's identity $\inner{\nabla_{W_k}f(\x;\thv)}{W_k}=d_kf(\x;\thv)$. Multiplying by
$\lambda_iy_i$ at $\x=\x_i$ and summing over $i$ gives \eqref{eq:euler}. Summing \eqref{eq:euler} over
$k$ gives $\inner{\rr(z)}{\thv}=L\,\Gamma(z)$. If $\rr(z)\in\Cc(\thv)$, write $\rr=t\kk$ with $t>0$
and $\kk\in\partial h(\thv)$; then $\inner{\rr}{\thv}=t\|\thv\|^2>0$ by
Lemma~\ref{prop:member}, so $\Gamma(z)>0$ and $\inner{\rr_{W_k}}{W_k}=d_k\Gamma(z)>0$ for every $k$,
which forces $\rr_{W_k}\neq\mathbf0$ and $W_k\neq\mathbf0$. For a solution of \eqref{eq:member},
$t=1$ and $\Gamma=\|\thv\|^2/L$.
\end{proof}

\begin{proof}[Proof of Corollary~\ref{cor:active}]
Let $z$ solve \eqref{eq:member}; by Lemma~\ref{lem:euler}, $\rr_{W_k}(z)\neq\mathbf0$ for all $k$.
(a) Lemma~\ref{lem:blockdecomp}(b) requires $\rr_{W_k}=\mathbf0$ for every block outside
$\argmax_j\|W_j\|_{\sms}/c_j$, so every block is in the arg-max. (b) By
Lemma~\ref{lem:subdiff}(ii), every element of $C(\thv)$ is supported on $A$; since
$\rr_{W_k}\neq\mathbf0$, block $k$ meets $A$, so $\|W_k\|_\infty=\|\thv\|_\infty$.
\end{proof}

\subsection{Exactness of per-block alignment}
\label{app:exact}

\begin{proof}[Proof of Proposition~\ref{prop:exact}]
The second equality holds for any $\thv$. A summand of \eqref{eq:align} vanishes iff
$\inner{\rr_{W_k}}{W_k}=\|\rr_{W_k}\|_{(k),\dstar}\|W_k\|_{(k)}$ with
$\rr_{W_k}\neq\mathbf0$, i.e.\ by \eqref{eq:aligndef} applied to the block norm, iff
$\rr_{W_k}/\|\rr_{W_k}\|_{(k),\dstar}\in\partial\|W_k\|_{(k)}$, i.e.\ iff
$\rr_{W_k}\in C_k(\thv)\setminus\{\mathbf0\}$. For Muon,
$\partial(\|\cdot\|_{\sms}/c_k)=c_k^{-1}\partial\|\cdot\|_{\sms}$, so $C_k$ and the summand do not
depend on $c_k$.

\emph{Max-type norms (Muon, Adam).} Write $\|\thv\|=\max_kN_k(W_k)$ with
$N_k=\|\cdot\|_{(k)}$. By Corollary~\ref{cor:active} all blocks attain the maximum, so the
subdifferential of a pointwise maximum of convex functions is the convex hull of the active
subdifferentials:
$\partial\|\thv\|=\conv\bigcup_k\{\iota_k\mathbf g:\mathbf g\in\partial N_k(W_k)\}$, with $\iota_k$
the embedding of block $k$. Since $\|\thv\|>0$,
$C(\thv)=\cone\partial\|\thv\|=\{\sum_kt_k\iota_k\mathbf g_k:t_k\ge0,\ \mathbf g_k\in\partial
N_k(W_k)\}=\prod_kC_k(\thv)$. Hence if $\rr\in\Cc(\thv)$ then
$\rr_{W_k}\in C_k(\thv)$ for all $k$, and $\rr_{W_k}\neq\mathbf0$ by Lemma~\ref{lem:euler}.
Conversely, if $\rr_{W_k}\in C_k(\thv)\setminus\{\mathbf0\}$ for all $k$, then
$\rr\in\prod_kC_k(\thv)=C(\thv)$ and $\rr\neq\mathbf0$.

\emph{$\ell_2$.} Here $C(\thv)=\{t\thv:t\ge0\}$ and $C_k(\thv)=\{tW_k:t\ge0\}$. Let $z^\star$ solve
\eqref{eq:member}, i.e.\ $\rr(z^\star)=\thv$; by Lemma~\ref{lem:euler},
$\|W_k\|_F^2=\inner{\rr_{W_k}(z^\star)}{W_k}=d_k\Gamma^\star$ with
$\Gamma^\star=\|\thv\|^2/L>0$. ($\subseteq$) If $\rr=t\thv$ with $t>0$, then
$\rr_{W_k}=tW_k\neq\mathbf0$. ($\supseteq$) If $\rr_{W_k}=t_kW_k$ with $t_k>0$ for all $k$, then
\eqref{eq:euler} gives $t_k\|W_k\|_F^2=d_k\Gamma(z)$, hence
$t_k=\Gamma(z)/\Gamma^\star$ for every $k$, and $\rr=t\thv$ with
$t=\Gamma(z)/\Gamma^\star>0$.
\end{proof}

\begin{proof}[Verification of Remark~\ref{rem:couple}]
On the zero set of $L_{\mathrm{align}}$, every block is tight, so
$\|\rr_{W_k}\|_{(k),\dstar}\|W_k\|_{(k)}=\inner{\rr_{W_k}}{W_k}=d_k\Gamma(z)$ by
\eqref{eq:euler}. Hence
$\log(\|\rr_{W_k}\|_{(k),\dstar}/\|W_k\|_{(k)})=\log\Gamma(z)+\log(d_k/\|W_k\|_{(k)}^2)$, and the
variance over $k$ removes the $k$-independent term $\log\Gamma(z)$.
\end{proof}

\subsection{Set-valued target and collapse}
\label{app:coupling}

\begin{proof}[Proof of Lemma~\ref{prop:scale}]
\emph{(a)} Numerator and denominator are positively $1$-homogeneous in each argument:
$\inner{a_k\rr_{W_k}}{b_kW_k}=a_kb_k\inner{\rr_{W_k}}{W_k}$,
$\|a_k\rr_{W_k}\|_{\nuc}=a_k\|\rr_{W_k}\|_{\nuc}$ and
$\|b_kW_k\|_{\sms}=b_k\|W_k\|_{\sms}$, so the ratio is unchanged.

\emph{(b)} By Corollary~\ref{cor:active} all blocks are active, and as in the proof of
Proposition~\ref{prop:exact},
$\partial\|\thv\|=\{(\mu_kc_k^{-1}G_k)_k:\mu\in\Delta_K,\ G_k\in\partial\|W_k\|_{\sms}\}$;
multiplying by $\|\thv\|$ gives $\partial h(\thv)$. On the relative interior ($\mu_k>0$) the
parametrization is injective, since $\|G_k\|_{\nuc}=\tr M=1$ recovers
$\mu_k=c_k\|\uu_{W_k}\|_{\nuc}/\|\thv\|$ from an element $\uu$ and then $G_k$. Each
$\partial\|W_k\|_{\sms}$ is a spectrahedron of dimension $\rho_k(\rho_k+1)/2-1$
(Lemma~\ref{lem:subdiff}), which gives the stated dimension. For a solution,
$\rr_{W_k}=\|\thv\|\mu_kc_k^{-1}G_k$, and since
$\inner{G_k}{W_k}=\sigma_1(W_k)=c_k\|\thv\|$, Lemma~\ref{lem:euler} gives
$\|\thv\|^2\mu_k=\inner{\rr_{W_k}}{W_k}=d_k\|\thv\|^2/L$, i.e.\ $\mu_k=d_k/L$. Hence
$\rr_{W_k}=\|\thv\|d_kG_k/(Lc_k)=d_k\sigma_1(W_k)G_k/(Lc_k^2)$, using
$\|\thv\|=\sigma_1(W_k)/c_k$. If $\rho_k=1$, $G_k=\pp_k\qq_k\T$ is unique, so
$\kk^\circ_{W_k}=(d_k/(Lc_k^2))\kk^\star_{W_k}$, which is a common positive multiple of $\kk^\star$
iff $d_k/c_k^2$ is constant. If $\rho_k>1$, $G_k$ ranges over a spectrahedron of positive dimension,
so stationarity does not select a unique point. The barycentric choice
$G_k=U_{1,k}V_{1,k}\T/\rho_k$ is basis-independent, but selecting it adds a condition not implied by
stationarity.

\emph{(c)} $\|\kk-t\sv\|^2=\|\kk\|^2-2t\inner{\kk}{\sv}+t^2\|\sv\|^2$ is a convex quadratic with
unconstrained minimizer $\inner{\kk}{\sv}/\|\sv\|^2$; projecting onto $t\ge0$ gives
$t^\star=(\inner{\kk}{\sv})_+/\|\sv\|^2$. If $\inner{\kk}{\sv}\le0$ then $t^\star=0$. Otherwise the
minimum is $\|\kk\|^2(1-\cos^2\angle(\kk,\sv))$, which vanishes iff $\sv$ is a positive multiple of
$\kk$. Each term of \eqref{eq:muon-loss} is invariant under $\rr\mapsto t\rr$, $t>0$, by (a).
\end{proof}

\subsection{Per-block versus global tightness for Muon}
\label{app:blockdecomp}

\begin{lemma}[Block decomposition of tightness]\label{lem:blockdecomp}
With $\|\thv\|=\max_k\|W_k\|_{\sms}/c_k$ and $\|\cdot\|_\dstar=\sum_kc_k\|\cdot\|_{\nuc}$, global
tightness $\inner{\rr}{\thv}=\|\rr\|_\dstar\|\thv\|$ holds if and only if
\emph{(a)} $\inner{\rr_{W_k}}{W_k}=\|\rr_{W_k}\|_{\nuc}\|W_k\|_{\sms}$ for every $k$ with
$\rr_{W_k}\neq0$, and \emph{(b)} $\rr_{W_k}=0$ unless
$k\in\argmax_j\|W_j\|_{\sms}/c_j$.
\end{lemma}
\begin{proof}
$\inner{\rr}{\thv}=\sum_k\inner{\rr_{W_k}}{W_k}\le\sum_k\|\rr_{W_k}\|_{\nuc}\|W_k\|_{\sms}
=\sum_k\big(c_k\|\rr_{W_k}\|_{\nuc}\big)\big(\|W_k\|_{\sms}/c_k\big)
\le\big(\sum_kc_k\|\rr_{W_k}\|_{\nuc}\big)\|\thv\|$. Equality in the first inequality is (a);
equality in the second requires
$c_k\|\rr_{W_k}\|_{\nuc}\big(\|\thv\|-\|W_k\|_{\sms}/c_k\big)=0$ for all $k$, which is (b).
\end{proof}

Condition (a) is what \eqref{eq:muon-loss} enforces. Condition (b) would be a severe additional
requirement if some block failed to attain the maximum, but by Corollary~\ref{cor:active} none does at
a KKT direction of a blockwise-homogeneous network, so (b) is vacuous there and \eqref{eq:muon-loss} is
exact (Proposition~\ref{prop:exact}).

\subsection{Specificity and collapse}
\label{app:proofs-count}

\begin{lemma}[Exact KKT targets]\label{lem:subdiff}
For $\thv\neq\mathbf0$: \emph{(i)} $\ell_2$:
$\partial\big(\tfrac12\|\thv\|_2^2\big)=\{\thv\}$;
\emph{(ii)} $\ell_\infty$:
$\partial\big(\tfrac12\|\thv\|_\infty^2\big)
=\|\thv\|_\infty\conv\{\sgn(\theta_j)\ee_j: j\in A\}$ with
$A=\argmax_l|\theta_l|$, a simplex of dimension $|A|-1$, whose cone is
$\{\rr:\rr_j=0\text{ for }j\notin A,\ \rr_j\theta_j\ge0\text{ for }j\in A\}$;
\emph{(iii)} spectral, for $W\neq0$ whose top singular value has multiplicity $\rho$:
$\partial\big(\tfrac12\|W\|_{\sms}^2\big)=\sigma_1\{U_1MV_1\T:M\succeq0,\ \tr M=1\}$, a
spectrahedron of dimension $\rho(\rho+1)/2-1$, which for $\rho=1$ is the singleton
$\{\sigma_1\pp\qq\T\}$.
\end{lemma}
\begin{proof}
By Lemma~\ref{lem:half}, $\partial(\tfrac12\|\cdot\|^2)=\|\cdot\|\,\partial\|\cdot\|$ in each case.
\emph{(i)} $\|\cdot\|_2$ is differentiable at $\thv\neq0$ with
$\partial\|\thv\|_2=\{\thv/\|\thv\|_2\}$. \emph{(ii)}
$\|\thv\|_\infty=\max_{j,\,\epsilon\in\{\pm1\}}\epsilon\,\theta_j$ is a maximum of linear functions,
so Danskin's theorem gives
$\partial\|\thv\|_\infty=\conv\{\sgn(\theta_j)\ee_j: j\in A\}$; nonnegative combinations of these
vectors are exactly the vectors supported on $A$ whose nonzero coordinates have matching signs,
equivalently $\rr_j\theta_j\ge0$ on $A$. \emph{(iii)} is the classical spectral-norm subdifferential
\citep{watson1992}; the map $M\mapsto U_1MV_1\T$ is injective, so the dimension is that of
$\{M\succeq0,\tr M=1\}$. For $\rho=1$, Danskin's theorem applied to
$\|W\|_{\sms}=\max_{\|\uu\|=\|\vv\|=1}\uu\T W\vv$ gives
$\partial\|W\|_{\sms}=\{\pp\qq\T\}$.
\end{proof}

\begin{proof}[Proof of Lemma~\ref{lem:specfloor}]
Since $\rho_k\le r_k$,
$m_kn_k-\rho_k(\rho_k+1)/2\ge m_kn_k-r_k(r_k+1)/2$, and
$m_kn_k-r_k(r_k+1)/2-\tfrac12(m_kn_k-r_k)=\tfrac12(m_kn_k-r_k^2)\ge0$. Sum over $k$.
\end{proof}

\begin{proof}[Proof of Lemma~\ref{lem:ceiling}]
For $f(\x;\thv)=\inner{\thv}{\phi(\x)}$,
$\nabla_{\thv}f(\x;\thv)=\phi(\x)$, so
$\rr(z)=\sum_i\lambda_iy_i\phi(\x_i)$ does not depend on $\thv$. The stationarity-feasible set
$\{z:\rr(z)\in\Cc(\thv)\}$ therefore depends on $\thv$ only through $C(\thv)$, and the global
alignment loss vanishes iff $\rr(z)\in\Cc(\thv)$ by \eqref{eq:aligndef}. Hence
$C(\thv)=C(\thv')$ implies that the feasible sets and zero sets coincide. Thus the dependence of this
stationarity-only formulation on the checkpoint factors through $C(\thv)$, whose image has dimension
$S$ on the stratum by Definition~\ref{def:spec}. No input non-identifiability conclusion follows from
this factorization alone.
\end{proof}

\begin{proof}[Proof of Lemma~\ref{lem:spec}]
\emph{$\ell_2$.} By Lemma~\ref{lem:subdiff}(i), $C(\thv)=\{t\thv:t\ge0\}$, which is determined exactly
by the direction $\thv/\|\thv\|_2\in S^{p-1}$, and distinct directions give distinct cones; hence
$S=p-1$.

\emph{Spectral.} With all blocks active, $C(\thv)=\prod_kC_k$ with
$C_k=\{U_{1,k}MV_{1,k}\T:M\succeq0\}$ (proof of Proposition~\ref{prop:exact} and
Lemma~\ref{lem:subdiff}(iii)); $\sigma_1(W_k)$ does not enter. $C_k$ determines the partial isometry
$P_k:=U_{1,k}V_{1,k}\T$: the elements of $C_k$ have singular values equal to the eigenvalues of $M$,
so up to positive scale $P_k$ is the unique element of rank $\rho_k$ with all nonzero singular values
equal ($M=cI$). Conversely, $P_k$ determines $C_k$, because any two factorizations of $P_k$ differ by
$(U_1,V_1)\mapsto(U_1Q,V_1Q)$ with $Q$ orthogonal, which leaves
$\{U_1MV_1\T:M\succeq0\}$ unchanged. The image of $\thv\mapsto C(\thv)$ on the stratum is therefore
the image of $\thv\mapsto(P_k)_k$. This image is open in the product of the manifolds $\mathcal P_k$
of rank-$\rho_k$ partial isometries: given $P_k'$ near $P_k$, write
$P_k'=Q_kP_kQ_k'^{\top}$ with orthogonal $Q_k,Q_k'$ near the identity, and replace $W_k$ by
$Q_kW_kQ_k'^{\top}$, which keeps all singular values, hence the stratum and activity of every block.
Finally, $\dim\mathcal P_k=S_k(\rho_k)$: the polar decomposition
$B=P\,(B\T B)^{1/2}$ identifies the manifold of rank-$\rho$ matrices, of dimension
$\rho(m+n-\rho)$, with pairs of a rank-$\rho$ partial isometry and a positive definite operator on its
$\rho$-dimensional row space, of dimension $\rho(\rho+1)/2$. For $\rho=1$,
$S_k=m_k+n_k-2$.

\emph{$\ell_\infty$.} By Lemma~\ref{lem:subdiff}(ii), $C(\thv)$ is a function of
$(A,\sgn\theta|_A)$ alone, which is discrete data; no continuous functional of $\thv$ is determined,
so $S=0$, for every $|A|=s$.
\end{proof}

\begin{proof}[Proof of the multiplicity bound in Appendix~\ref{app:specremark}]
$S_k(\rho)=\rho(m+n)-\tfrac32\rho^2-\tfrac12\rho$ is concave, so on $[1,r]$ it is at least
$\min\{S_k(1),S_k(r)\}$. With $r=n\le m$,
$S_k(r)=nm-n(n+1)/2\ge n(n-1)/2$, and
$S_k(r)-(m+n-2)=m(n-1)-n(n+3)/2+2\ge n(n-1)-n(n+3)/2+2=(n-1)(n-4)/2\ge0$ for $n\ge4$.
\end{proof}

\begin{proof}[Proof of Proposition~\ref{prop:nonid}]
By Lemma~\ref{lem:subdiff}(ii), $C(\thv)$ depends only on $(A,\sgn\thv|_A)$. The conclusion about
stationarity-feasible sets follows from Lemma~\ref{lem:ceiling}. The final qualification is necessary:
even a cone that is constant on a stratum can impose equations on $\rr(z)$ that identify $z$ for a
particular feature map.
\end{proof}

\begin{remark}[Nondegeneracy for linear-in-parameter models]\label{rem:lp}
For $f=\inner{\thv}{\phi(\x)}$, the $\ell_\infty$ max-margin problem is the linear program
$\min t$ s.t.\ $t-\theta_j\ge0$, $t+\theta_j\ge0$,
$y_i\inner{\thv}{\phi(\x_i)}\ge1$, in the variables $(\thv,t)\in\R^{p+1}$. If its solution is unique it
is a vertex, so the normals of the active constraints, namely
$(-\sgn(\theta_j)\ee_j,1)$ for $j\in A$ and $(y_i\phi(\x_i),0)$ for the active margins, span
$\R^{p+1}$, and their $\thv$-components span $\R^p$. Under strict complementarity the active margins
are exactly $I$, which is the span condition of Assumption~\ref{ass:nondeg}.
\end{remark}

\begin{proof}[Proof of Theorem~\ref{thm:linf}]
\emph{(i)} The spanning set of Assumption~\ref{ass:nondeg} has at most $s+n_I$ elements, so
$p\le s+n_I$. By Lemma~\ref{lem:subdiff}(ii), $C(\thv)$ is a full-dimensional cone in
$\operatorname{span}\{\ee_j:j\in A\}$, so $\codim C(\thv)=p-s$.

\emph{(ii)} $\aff C(\thv)=\operatorname{span}\{\ee_j:j\in A\}$, so $P_N$ is the coordinate projection
onto the $p-s$ coordinates outside $A$, and $\Phi$ takes values in $\R^{p-s}$. Hence
$\rank D\Phi(z_I^\star)\le p-s\le n_I$. If $n_Id\ge2$ then
$n_I<n_I(d+1)-1=\dim z_I-\dim\mathcal G$, so the rank condition of Definition~\ref{def:jac} fails.

\emph{(iii)} By the constant-rank theorem, $\Phi^{-1}(\mathbf0)$ is, near $z_I^\star$, a $C^1$
manifold $\mathcal M$ of dimension $n_I(d+1)-\rank D\Phi\ge n_Id$, and it contains $z_I^\star$
because $\rr(z^\star)\in C(\thv)$. On $\mathcal M$ the coordinates of $\rr$ outside $A$ vanish. The
coordinates $\rr_j$, $j\in A$, are continuous and nonzero with sign $\sgn\theta_j$ at $z^\star$, hence
keep that sign nearby, and $\lambda_i>0$ persists nearby; so $\rr(z)\in\Cc(\thv)$ on a neighbourhood of
$z_I^\star$ in $\mathcal M$. For the projection $\pi$ onto the input coordinates,
$\ker(D\pi|_{T\mathcal M})$ lies in the $n_I$ multiplier coordinates, so
$\rank D(\pi|_{\mathcal M})\ge n_Id-n_I$ everywhere. Its maximal value is attained on an open set
accumulating at $z_I^\star$, and there the constant-rank theorem makes the image a $C^1$ manifold of
dimension at least $n_I(d-1)$.
\end{proof}

\begin{remark}[Codimensions in \eqref{eq:codim}]
The cone $C$ has dimension $1$ for $\ell_2$; $\sum_k\rho_k(\rho_k+1)/2$ for the spectral norm (a
product of per-block spectrahedral cones, all blocks active), which is $K$ when all $\rho_k=1$; and $s$
for $\ell_\infty$. The codimensions in \eqref{eq:codim} follow.
\end{remark}

\section{Local Identifiability}
\label{app:local}

Specificity measures the continuous variability of the optimizer-dependent target cone
(Lemma~\ref{lem:ceiling}). The complementary question, whether the true data is regularly isolated
among stationarity-consistent candidates, is one of well-posedness in the attacker's own variables.
Only candidates with positive multipliers enter the local residual analysis, so we restrict to them.

\begin{definition}[Regular first-order identifiability]\label{def:jac}
Let $z^\star$ be the true solution, $I=\{i:\lambda_i^\star>0\}$, $n_I=|I|$, and
$z_I=(\{\x_i\},\{\lambda_i\})_{i\in I}\in\R^{n_I(d+1)}$, the remaining candidates being held at
$z^\star$. Let $P_{N(\thv)}$ be the orthogonal projection onto the orthogonal complement of
$\aff C(\thv)$ and $J=\partial_{z_I}\big[P_{N(\thv)}\rr(z)\big]$ at $z^\star$. Call $z^\star$
\emph{regularly first-order identifiable} if
$\rank J=\dim z_I-\dim\mathcal G=n_I(d+1)-1$, where $\mathcal G$ is the one-dimensional multiplier
gauge $\lambda\mapsto c\lambda$ of the cone formulation \eqref{eq:cone}. Discrete symmetries
(permutations of candidates) do not affect the rank.
\end{definition}

The multiplier gauge always lies in $\ker J$: along $(\mathbf0,\lambda_I^\star)$ the derivative of
$P_{N(\thv)}\rr$ is $P_{N(\thv)}\rr(z^\star)=\mathbf0$. Maximal rank rules out every infinitesimal
ambiguity except this gauge and is a sufficient regularity condition for local isolation modulo gauge.
Its failure does not, by itself, prove set-theoretic non-identifiability, because a rank-deficient
nonlinear equation can still have an isolated zero.

\begin{lemma}[A necessary condition]\label{lem:Bnec}
$\rank J\le\codim C(\thv)$. Hence regular first-order identifiability requires
$\codim C(\thv)\ge n_I(d+1)-1$.
\end{lemma}
\begin{proof}
$C(\thv)$ is a cone containing $\mathbf0$, so $\aff C(\thv)=\operatorname{span}C(\thv)$ has dimension
$\dim C(\thv)$, and $P_{N(\thv)}\rr$ takes values in a space of dimension
$p-\dim C(\thv)=\codim C(\thv)$.
\end{proof}

By \eqref{eq:codim} and Lemma~\ref{lem:specfloor}, the necessary condition holds for $\ell_2$ and for
the spectral norm whenever $p-1$, respectively $\tfrac12(p-\sum_kr_k)$, is at least
$n_I(d+1)-1$, which is the over-parameterized regime. For $\ell_\infty$ it fails at nondegenerate KKT
points:

\begin{proposition}[Failure of regular first-order identifiability]\label{prop:Bneg}
Under Assumption~\ref{ass:nondeg} and $n_Id\ge2$, $z^\star$ is not regularly first-order identifiable
for $\ell_\infty$, at any width. Under the additional hypotheses of Theorem~\ref{thm:linf}(iii), it
belongs to a positive-dimensional local family of stationarity-consistent candidates.
\end{proposition}
\begin{proof}
Theorem~\ref{thm:linf}(ii)--(iii).
\end{proof}

\begin{conjecture}\label{conj:jac}
For $\ell_2$, and for the spectral norm, $z^\star$ is regularly first-order identifiable for
sufficiently over-parameterized blockwise-homogeneous $f$.
\end{conjecture}

We prove Adam's failure of regular first-order identifiability (Proposition~\ref{prop:Bneg}) and, under
the additional hypotheses of Theorem~\ref{thm:linf}(iii), actual positive-dimensional local ambiguity.
For the positive $\ell_2$ and spectral direction we prove only the necessary condition of
Lemma~\ref{lem:Bnec}. The conjectured positive half says that high specificity and high codimension are
not merely necessary but sufficient for regular first-order well-posedness. It is directly measurable,
requiring one singular value decomposition of $J$ at a recovered solution.

\thispagestyle{empty}

\begin{figure*}[htb]
\centering
\begin{subfigure}[b]{0.32\textwidth}
  \centering
  \includegraphics[width=\linewidth]{mnist_muon.png}
  \caption{Muon}
\end{subfigure}\hfill
\begin{subfigure}[b]{0.32\textwidth}
  \centering
  \includegraphics[width=\linewidth]{mnist_sgd.png}
  \caption{SGD}
\end{subfigure}\hfill
\begin{subfigure}[b]{0.32\textwidth}
  \centering
  \includegraphics[width=\linewidth]{mnist_adam.png}
  \caption{Adam}
\end{subfigure}

\vspace{0.4em}
\begin{subfigure}[b]{0.32\textwidth}
  \centering
  \includegraphics[width=\linewidth]{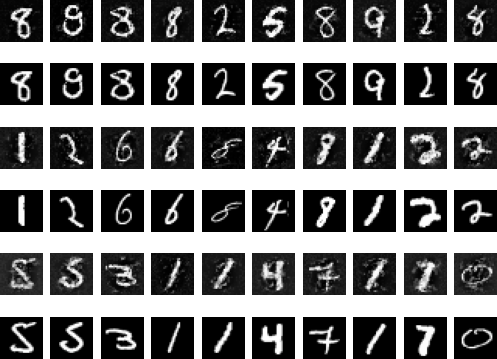}
  \caption{Muon, range prior}
\end{subfigure}\hfill
\begin{subfigure}[b]{0.32\textwidth}
  \centering
  \includegraphics[width=\linewidth]{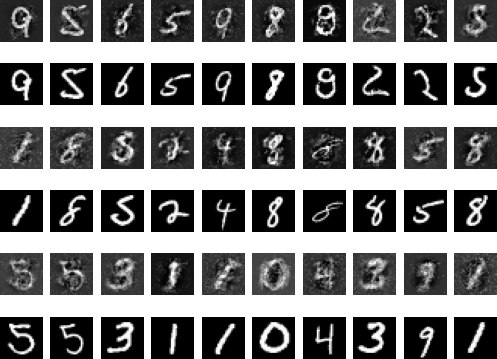}
  \caption{SGD, range prior}
\end{subfigure}\hfill
\begin{subfigure}[b]{0.32\textwidth}
  \centering
  \includegraphics[width=\linewidth]{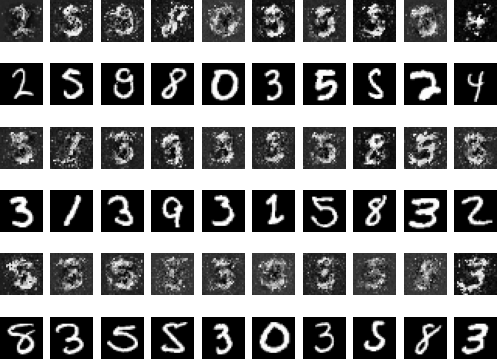}
  \caption{Adam, range prior}
\end{subfigure}

\caption{Reconstructions on binary MNIST ($n=500$) without a prior (top) and with the range prior
(bottom). In each panel, odd rows show reconstructions and the row below shows the matched training
images. Muon recovers sharp digit structure even without a prior.}
\label{fig:qual-mnist}
\end{figure*}

\section{Experimental Details}
\label{app:details}

\paragraph{Multiclass residual.} For multiclass models the residual uses one multiplier per sample and
competing class, $\sum_i\sum_{j\neq y_i}\lambda_{ij}\nabla_{\thv}[\Phi_{y_i}(\x_i;\thv)-\Phi_j(\x_i;\thv)]$
\citep{buzaglo2023}.

\paragraph{Test data.} For 10-class CIFAR-10 we evaluate on a class-balanced test subset of
$1{,}000$ images ($100$ per class).

\paragraph{Initialization.} Following \citet{haim2022}, the
first layer is initialized with variance $10^{-4}$ and the remaining layers with Kaiming normal
initialization \citep{he2015delving}.

\paragraph{Trained models \& hardware.}
Table~\ref{tab:train} summarizes the trained model configurations used in Section~5. All networks were trained using PyTorch on a single NVIDIA GeForce GTX 1080 Ti GPU. Training required fewer than $2$ minutes for $10,000$ iterations over \(N=500\) training samples with learning rate $10^{-2}$. For the binary classification setup, $250$ samples per class were used; for the $10$-class multiclass setup, $50$ samples per class were used. Code and reproduction scripts will be made publicly available.

\begin{table}[htb]
\centering
\caption{Trained models ($n=500$). All models interpolate the training set. Because the network is
homogeneous, an unnormalized margin depends on the parameter scale and is not comparable across
optimizers.}
\label{tab:train}
\small
\resizebox{\linewidth}{!}{
\begin{tabular}{llcccc}
\toprule
Task & Optimizer & Test acc.\ (\%) & Test loss & Min.\ margin\\
\midrule
\multirow{3}{*}{MNIST (binary)}
 & SGD      & $90.0$ & $0.790$ & $9.21$\\
 & Adam     & $91.0$ & $0.753$ & $16.64$\\
 & Muon & $92.0$ & $0.853$ & $16.65$\\
\midrule
\multirow{3}{*}{CIFAR-10 (binary)}
 & SGD      & $77.0$ & $2.25$ & $9.56$\\
 & Adam     & $79.0$ & $3.41$ & $16.64$\\
 & Muon & $79.0$ & $2.57$ & $16.63$\\
\midrule
\multirow{3}{*}{CIFAR-10 (10-class)}
 & SGD      & $28.0$ & $7.97$ & $13.75$\\
 & Adam     & $26.0$ & $13.5$ & $14.11$\\
 & Muon & $29.0$ & $7.09$ & $15.46$\\
\bottomrule
\end{tabular}
}
\end{table}

\paragraph{Muon geometry.}
In our setup, Muon is applied exclusively to internal 2D weight matrices (e.g., hidden-to-hidden layer \(W_2\)), whereas standard SGD \citep{battash2024revisiting} is used for both the input layer (\(W_1\)) and the final output layer (\(W_L\)). Consequently, the overall implicit bias is spectral norm, with nuclear dual norm on the internal blocks and Euclidean (\(L_2\)) on the first and last layers.

\paragraph{Data reconstruction attack implementation.}
Data reconstruction was executed by solving the KKT stationarity conditions corresponding to SGD, Adam, and Muon. Specifically, for layers optimized using SGD, the stationarity loss is defined by the stationarity term in Equation~\ref{eq:haim}; for layers utilizing the Adam optimizer, the stationarity loss follows Equation~\ref{eq:adam_stats_loss}; and for Muon optimizer layers, it is given by Equation~\ref{eq:muon-loss}.

To facilitate gradient-based optimization during reconstruction without altering forward behavior, we adopted a modified backward pass for the ReLU activation function \(y = \max(0, x)\) following \citet{haim2022}. Specifically, while the forward pass remains exact ReLU, the backward gradient is modified as:
\begin{equation}
  \widetilde{\frac{\partial L}{\partial x}} =\frac{\partial L}{\partial y}\,\sigma(\alpha x), \quad \text{where } \sigma(z) = \frac{1}{1 + e^{-z}}
\end{equation}
Here, \(\alpha\) acts as a hyperparameter interpolating between a scaled linear backward pass function (\(\alpha \to 0\)) and the standard ReLU derivative (\(\alpha \to \infty\)). 

Dual variables (Lagrange multipliers associated with stationarity conditions) and reconstructed input samples \(x_i\) were optimized. For the attack optimization (applied across all three attacks), we followed \citet{haim2022} using SGD with a momentum of $0.9$ over $100,000$ iterations.

\paragraph{Hyperparameter search.}
We selected reconstruction hyperparameters through a random search maximizing SSIM@30:
\begin{itemize}
    \item \textbf{Reconstruction learning rate:} Log-uniform sampling in \([10^{-4}, 10^{-2}]\).
    \item \textbf{Initialization scale (\(\sigma_x\)):} Log-uniform sampling in \([10^{-3}, 1]\) for the initial scale of \(x_i\).
    \item \textbf{Modified ReLU parameter (\(\alpha\)):} Uniform sampling in \([10, 500]\).
\end{itemize}

\begin{figure*}[t]
\centering
\begin{subfigure}[b]{0.32\textwidth}
  \centering
  \includegraphics[width=\linewidth]{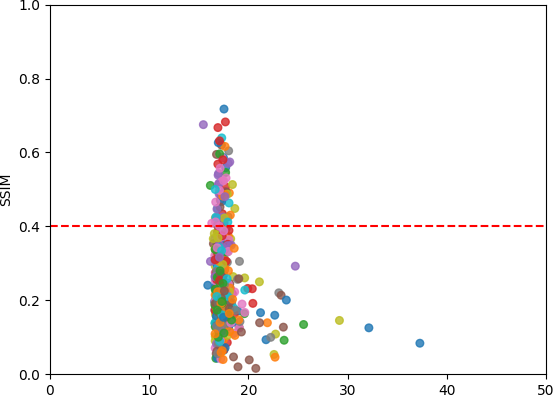}
  \caption{Muon, 10-class}
\end{subfigure}\hfill
\begin{subfigure}[b]{0.32\textwidth}
  \centering
  \includegraphics[width=\linewidth]{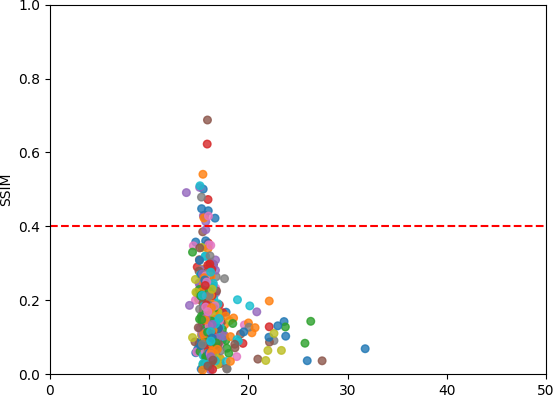}
  \caption{SGD, 10-class}
\end{subfigure}\hfill
\begin{subfigure}[b]{0.32\textwidth}
  \centering
  \includegraphics[width=\linewidth]{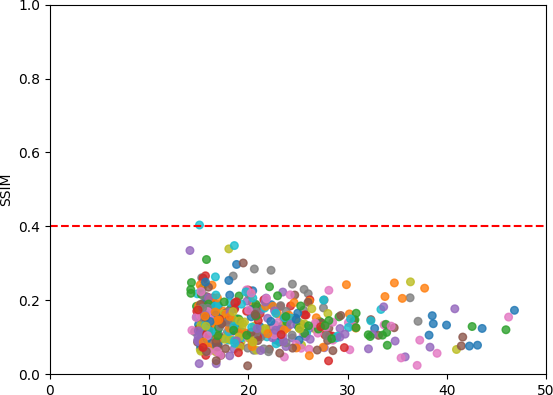}
  \caption{Adam, 10-class}
\end{subfigure}

\caption{Per-sample SSIM versus training margin $\gamma_i$ on CIFAR-10 ($n=500$, range prior), for
multiclass models. Each point is a training sample, colored by class; the
dashed line marks $\mathrm{SSIM}=0.4$. Recognizable reconstructions occur only near the minimal
margin. Muon produces the most of them and Adam essentially none.}
\label{fig:ssim-margin-multi}
\end{figure*}

\paragraph{Post-processing \& matching.}
\label{app:hung}
To evaluate fidelity, reconstructed samples were matched one-to-one to ground-truth training samples using the Hungarian algorithm on centered, normalized $\ell_2$ distance. The best-matched instances (such as the top 30 pairs) were selected for qualitative and quantitative reporting.

\section{Additional Experimental Results}
\label{app:extra}

To complement the qualitative visual comparisons presented in the main text, Tables~\ref{tab:full-mnist}--\ref{tab:full-multi} report the complete quantitative reconstruction results across all experimental configurations. We evaluate performance across both binary and multiclass tasks on MNIST and CIFAR-10 under multiple prior regimes: without prior constraints, under range prior constraints \([0, 1]\), and with additional regularized variants including Total Variation (TV) and \(L_1\) sparsity priors. Reconstruction fidelity is evaluated using standard pixel-level and structural metrics, specifically Peak Signal-to-Noise Ratio (PSNR), Structural Similarity Index (SSIM), \(L_1\) distance, and \(L_2\) distance.

Across all datasets, and prior configurations, \mbox{Muon} consistently achieves the superior value in every single metric entry, outperforming both standard SGD and Adam. Remarkably, even in the completely unconstrained setting (without prior), \mbox{Muon} retains significantly higher structural fidelity, highlighting the high specificity of reconstructions driven by its implicit bias.

\paragraph{Multiclass training dynamics.} 
While the overall performance ordering (\(\text{Muon} > \text{SGD} > \text{Adam}\)) is largely maintained in the $10$-class setting, the quantitative gap between Muon and SGD narrows compared to the binary regime. Specifically, Muon exhibits a higher concentration of samples exceeding the $0.4$ recognizability threshold compared to SGD, as illustrated in Figure~\ref{fig:ssim-margin-multi}.

\paragraph{Comparing across sample sizes (Table~\ref{tab:abl}).} 
Evaluating reconstruction fidelity across varying dataset sizes $n$ reveals that SSIM@30 peaks at $n=100$ and decreases substantially for $n \ge 500$. Although selecting the top 30 instances from a larger candidate pool ($m = 2n$) represents a more selective subset, the observed drop in performance indicates that the underlying inverse problem becomes fundamentally harder to invert accurately as the scale $n$ increases.

\paragraph{Computational complexity.}
Let \(q\) be the number of reconstruction candidates, \(p\) the number of model parameters, and \(d\) the input dimension. Each attack iteration performs work equivalent to \(q\) differentiable parameter-gradient evaluations plus \(O(p)\) residual aggregation. The SGD and Adam alignment terms require \(O(p)\) additional work, while Muon's dual-norm term additionally requires blockwise singular-value computations. Excluding model activations and automatic-differentiation workspace, the explicit attack state requires \(O(qd+p)\) memory. We study the empirical dependence on training-set size in Section~\ref{sec:exp} and Table~\ref{tab:abl}.

\newcommand{\fullheader}{
\toprule
Prior & Optimizer & SSIM$\uparrow$ & SSIM@30$\uparrow$ & PSNR$\uparrow$ & PSNR@30$\uparrow$ & $\ell_1\downarrow$ & $\ell_1$@30$\downarrow$ & $\ell_2\downarrow$ & $\ell_2$@30$\downarrow$\\
\midrule}

\begin{figure*}[htb]
\centering
\begin{subfigure}[b]{0.32\textwidth}
  \centering
  \includegraphics[width=\linewidth]{cifar10_muon.png}
  \caption{Muon}
\end{subfigure}\hfill
\begin{subfigure}[b]{0.32\textwidth}
  \centering
  \includegraphics[width=\linewidth]{cifar10_sgd.png}
  \caption{SGD}
\end{subfigure}\hfill
\begin{subfigure}[b]{0.32\textwidth}
  \centering
  \includegraphics[width=\linewidth]{cifar10_adam.png}
  \caption{Adam}
\end{subfigure}

\vspace{0.4em}
\begin{subfigure}[b]{0.32\textwidth}
  \centering
  \includegraphics[width=\linewidth]{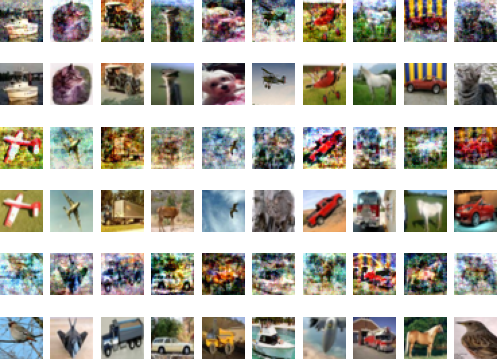}
  \caption{Muon, range prior}
\end{subfigure}\hfill
\begin{subfigure}[b]{0.32\textwidth}
  \centering
  \includegraphics[width=\linewidth]{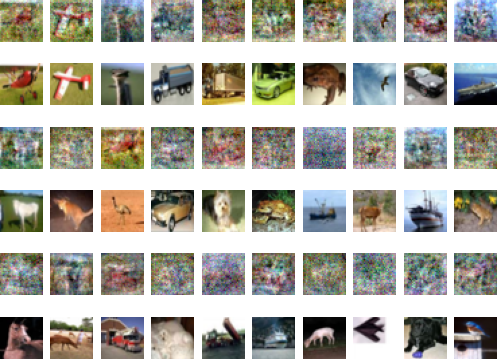}
  \caption{SGD, range prior}
\end{subfigure}\hfill
\begin{subfigure}[b]{0.32\textwidth}
  \centering
  \includegraphics[width=\linewidth]{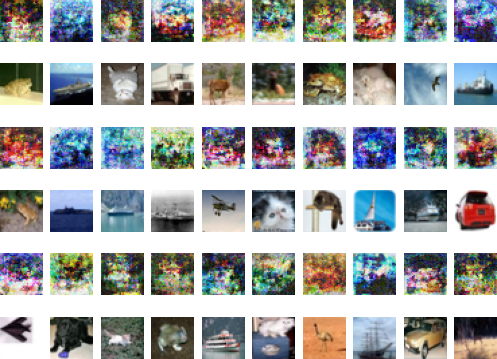}
  \caption{Adam, range prior}
\end{subfigure}

\caption{Reconstructions on binary CIFAR-10 ($n=500$), in the layout of Figure~\ref{fig:qual-mnist}.
Muon reconstructions best preserve object shape and color.}
\label{fig:qual-cifar}
\end{figure*}

\renewcommand{\B}[1]{\textbf{#1}}

\begin{table}[htb]
\centering
\caption{Reconstruction on binary MNIST ($n=500$). Best per prior in \textbf{bold}.}
\label{tab:full-mnist}
\small
\resizebox{\linewidth}{!}{
\begin{tabular}{llcccccccc}
\fullheader
\multirow{3}{*}{None}
 & SGD      & 0.1661 & 0.3463 & 10.857 & 14.867 & 0.2059 & 0.1260 & 0.2865 & 0.1806\\
 & Adam     & 0.1527 & 0.2487 & 10.345 & 10.903 & 0.2077 & 0.1926 & 0.3039 & 0.2850\\
 & Muon & \B{0.2350} & \B{0.4310} & \B{11.495} & \B{16.997} & \B{0.1718} & \B{0.0935} & \B{0.2662} & \B{0.1413}\\
\midrule
\multirow{3}{*}{Range}
 & SGD      & 0.2007 & 0.4057 & 11.186 & 15.464 & 0.1700 & 0.1008 & 0.2759 & 0.1686\\
 & Adam     & 0.1898 & 0.2994 & 10.640 & 11.933 & 0.1958 & 0.1415 & 0.2938 & 0.2531\\
 & Muon & \B{0.3242} & \B{0.6371} & \B{12.115} & \B{20.154} & \B{0.1339} & \B{0.0517} & \B{0.2479} & \B{0.0982}\\
\midrule
\multirow{3}{*}{TV}
 & SGD      & 0.1811 & 0.3768 & 11.042 & 14.948 & 0.1964 & 0.1209 & 0.2805 & 0.1789\\
 & Adam     & 0.1739 & 0.2882 & 10.593 & 11.270 & 0.1986 & 0.1801 & 0.2954 & 0.2732\\
 & Muon & \B{0.3288} & \B{0.5808} & \B{11.781} & \B{17.699} & \B{0.1550} & \B{0.0783} & \B{0.2576} & \B{0.1303}\\
\midrule
\multirow{3}{*}{$\ell_1$}
 & SGD      & 0.1688 & 0.3614 & 10.894 & 14.845 & 0.2007 & 0.1227 & 0.2853 & 0.1810\\
 & Adam     & 0.1445 & 0.1752 & 10.340 & 11.985 & 0.1829 & 0.1433 & 0.3041 & 0.2516\\
 & Muon & \B{0.2849} & \B{0.5302} & \B{11.610} & \B{17.291} & \B{0.1563} & \B{0.0819} & \B{0.2627} & \B{0.1366}\\
\midrule
\multirow{3}{*}{TV$+\ell_1$}
 & SGD      & 0.1848 & 0.3977 & 11.086 & 15.423 & 0.1908 & 0.1120 & 0.2791 & 0.1694\\
 & Adam     & 0.1606 & 0.2311 & 10.538 & 12.028 & 0.1825 & 0.1487 & 0.2972 & 0.2504\\
 & Muon & \B{0.3299} & \B{0.6211} & \B{11.831} & \B{18.002} & \B{0.1489} & \B{0.0728} & \B{0.2561} & \B{0.1259}\\
\midrule
\multirow{3}{*}{All}
 & SGD      & 0.2289 & 0.4511 & 11.345 & 15.657 & 0.1641 & 0.0961 & 0.2709 & 0.1649\\
 & Adam     & 0.1758 & 0.2655 & 10.834 & 13.006 & 0.1663 & 0.1215 & 0.2873 & 0.2237\\
 & Muon & \B{0.3400} & \B{0.6599} & \B{11.996} & \B{20.126} & \B{0.1361} & \B{0.0520} & \B{0.2513} & \B{0.0986}\\
\bottomrule
\end{tabular}}
\end{table}

\begin{table}[htb]
\centering
\caption{Reconstruction on binary CIFAR-10 ($n=500$). Best per prior in \textbf{bold}.}
\label{tab:full-cifar}
\small
\resizebox{\linewidth}{!}{
\begin{tabular}{llcccccccc}
\fullheader
\multirow{3}{*}{None}
 & SGD      & 0.0986 & 0.2287 & 11.671 & 13.523 & 0.2027 & 0.1623 & 0.2609 & 0.2108\\
 & Adam     & 0.1365 & 0.1379 & 12.422 & 12.902 & 0.1849 & 0.1743 & 0.2393 & 0.2264\\
 & Muon & \B{0.1771} & \B{0.5508} & \B{12.567} & \B{17.802} & \B{0.1807} & \B{0.0974} & \B{0.2353} & \B{0.1288}\\
\midrule
\multirow{3}{*}{Range}
 & SGD      & 0.0994 & 0.2336 & 11.665 & 13.435 & 0.2027 & 0.1634 & 0.2611 & 0.2129\\
 & Adam     & 0.1299 & 0.1367 & 12.334 & 13.441 & 0.1863 & 0.1624 & 0.2417 & 0.2128\\
 & Muon & \B{0.1921} & \B{0.6130} & \B{12.808} & \B{19.190} & \B{0.1750} & \B{0.0825} & \B{0.2289} & \B{0.1098}\\
\midrule
\multirow{3}{*}{TV}
 & SGD      & 0.1106 & 0.2574 & 11.857 & 14.099 & 0.1975 & 0.1517 & 0.2554 & 0.1973\\
 & Adam     & 0.0958 & 0.2115 & 11.2531 & 13.5265 & 0.2122 & 0.1789 & 0.2656 & 0.2001\\
 & Muon & \B{0.2310} & \B{0.6132} & \B{13.202} & \B{18.974} & \B{0.1672} & \B{0.0852} & \B{0.2187} & \B{0.1125}\\
\midrule
\multirow{3}{*}{$\ell_1$}
 & SGD      & 0.0984 & 0.2375 & 11.668 & 13.995 & 0.2026 & 0.1528 & 0.2610 & 0.1996\\
 & Adam     & 0.0977 & 0.2112 & 10.2658 & 13.2568 & 0.2471 & 0.1793 & 0.2787 & 0.2323\\
 & Muon & \B{0.1933} & \B{0.5598} & \B{12.798} & \B{18.109} & \B{0.1750} & \B{0.0936} & \B{0.2291} & \B{0.1243}\\
\midrule
\multirow{3}{*}{TV$+\ell_1$}
 & SGD      & 0.1113 & 0.2440 & 11.872 & 13.748 & 0.1973 & 0.1581 & 0.2549 & 0.2054\\
 & Adam     & 0.1218 & 0.2369 & 11.2451 & 13.1178 & 0.2284 & 0.1955 & 0.2854 & 0.2323\\
 & Muon & \B{0.2305} & \B{0.6286} & \B{13.216} & \B{19.185} & \B{0.1667} & \B{0.0832} & \B{0.2184} & \B{0.1098}\\
\midrule
\multirow{3}{*}{All}
 & SGD      & 0.1122 & 0.2536 & 11.865 & 14.194 & 0.1975 & 0.1502 & 0.2551 & 0.1951\\
 & Adam     & 0.1827 & 0.2377 & 12.727 & 14.462 & 0.1768 & 0.1407 & 0.2310 & 0.1892\\
 & Muon & \B{0.2196} & \B{0.6189} & \B{13.138} & \B{18.848} & \B{0.1683} & \B{0.0852} & \B{0.2203} & \B{0.1142}\\
\bottomrule
\end{tabular}}
\end{table}

\begin{table}[htb]
\centering
\caption{Reconstruction on 10-class CIFAR-10 ($n=500$). Best per prior in \textbf{bold}.}
\label{tab:full-multi}
\small
\resizebox{\linewidth}{!}{
\begin{tabular}{llcccccccc}
\fullheader
\multirow{3}{*}{None}
 & SGD      & 0.2031 & 0.3827 & 11.264 & 13.435 & 0.1912 & 0.1657 & 0.2535 & 0.1906\\
 & Adam     & 0.1530 & 0.2066 & 12.254 & 12.740 & 0.1856 & 0.1758 & 0.2440 & 0.2307\\
 & Muon & \B{0.2446} & \B{0.5339} & \B{13.284} & \B{18.826} & \B{0.1651} & \B{0.0884} & \B{0.2167} & \B{0.1145}\\
\midrule
\multirow{3}{*}{Range}
 & SGD      & 0.2157 & 0.4515 & 12.999 & 16.445 & 0.1722 & 0.1158 & 0.2239 & 0.1506\\
 & Adam     & 0.1423 & 0.1771 & 12.250 & 13.238 & 0.1874 & 0.1653 & 0.2441 & 0.2178\\
 & Muon & \B{0.2445} & \B{0.5315} & \B{13.278} & \B{18.987} & \B{0.1651} & \B{0.0863} & \B{0.2168} & \B{0.1124}\\
\midrule
\multirow{3}{*}{TV}
 & SGD      & 0.2112 & 0.4332 & 12.977 & 16.134 & 0.1729 & 0.1198 & 0.2245 & 0.1561\\
 & Adam     & 0.2310 & 0.3442 & 13.520 & 15.774 & 0.1603 & 0.1224 & 0.2109 & 0.1627\\
 & Muon & \B{0.2761} & \B{0.5481} & \B{13.880} & \B{19.199} & \B{0.1537} & \B{0.0848} & \B{0.2023} & \B{0.1097}\\
\midrule
\multirow{3}{*}{$\ell_1$}
 & SGD      & 0.2081 & 0.4141 & 12.952 & 16.344 & 0.1731 & 0.1147 & 0.2251 & 0.1523\\
 & Adam     & 0.1382 & 0.2739 & 11.940 & 14.062 & 0.1913 & 0.1491 & 0.2529 & 0.1981\\
 & Muon & \B{0.2529} & \B{0.5371} & \B{13.436} & \B{18.686} & \B{0.1615} & \B{0.0900} & \B{0.2129} & \B{0.1163}\\
\midrule
\multirow{3}{*}{TV$+\ell_1$}
 & SGD      & 0.2110 & 0.4343 & 13.009 & 16.085 & 0.1721 & 0.1197 & 0.2236 & 0.1569\\
 & Adam     & 0.1946 & 0.3050 & 12.742 & 14.106 & 0.1756 & 0.1499 & 0.2306 & 0.1971\\
 & Muon & \B{0.2806} & \B{0.5890} & \B{13.791} & \B{19.408} & \B{0.1547} & \B{0.0821} & \B{0.2044} & \B{0.1071}\\
\midrule
\multirow{3}{*}{All}
 & SGD      & 0.2110 & 0.4343 & 13.009 & 16.085 & 0.1721 & 0.1197 & 0.2236 & 0.1569\\
 & Adam     & 0.1598 & 0.2603 & 12.491 & 13.821 & 0.1809 & 0.1571 & 0.2374 & 0.2037\\
 & Muon & \B{0.2814} & \B{0.5795} & \B{13.780} & \B{19.279} & \B{0.1549} & \B{0.0837} & \B{0.2046} & \B{0.1087}\\
\bottomrule
\end{tabular}}
\end{table}

\begin{figure*}[t]
  \centering
  \begin{subfigure}[b]{0.48\textwidth}
    \centering
    \includegraphics[width=\linewidth]{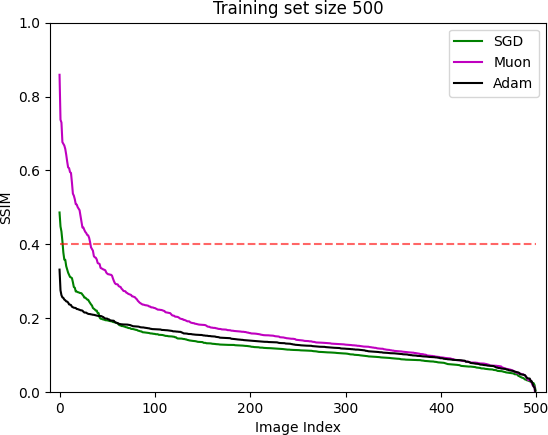}
    \caption{CIFAR-10 (\(N=500\), No Prior)}
    \label{fig:ssim_sorted_cifar10}
  \end{subfigure}
  \hfill
  \begin{subfigure}[b]{0.48\textwidth}
    \centering
    \includegraphics[width=\linewidth]{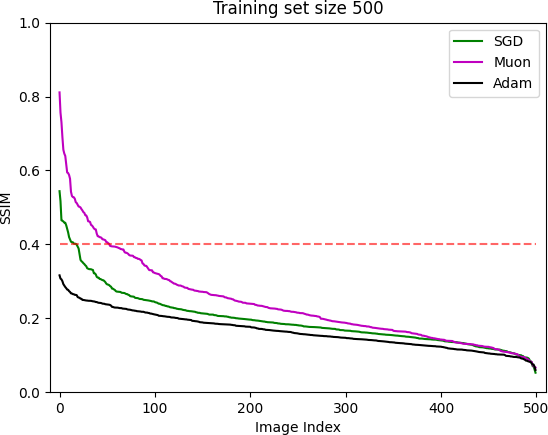}
    \caption{MNIST (\(N=500\), No Prior)}
    \label{fig:ssim_sorted_mnist}
  \end{subfigure}

  \caption{Sorted image-level SSIM values (from highest to lowest) for reconstructed samples on CIFAR-10 and MNIST without prior constraints (\(n=500\)). The dashed horizontal reference line at \(\text{SSIM} = 0.4\) indicates a threshold for acceptable visual reconstruction quality. Reconstructions under Muon consistently achieve higher SSIM scores, resulting in a substantially larger fraction of images reconstructed above $0.4$ compared to SGD and Adam (\(\text{Muon} > \text{SGD} > \text{Adam}\)).}
  \label{fig:ssim_sorted_comparison}
\end{figure*}

\section{Further Material}
\label{app:main-moved}

\subsection{Notation details (from \S\ref{sec:notation})}
\label{app:notation}

For nonzero $\uu$ and $\thv$, alignment means that $\uu$ is a positive multiple of $\thv$ in the $\ell_2$ geometry. For a single spectral-norm block, it means that $\uu=U_1 M V_1\T$ for some positive semidefinite $M$, where $U_1$ and $V_1$ span the corresponding top singular subspaces of $\thv$. For the max-spectral norm, aligned blocks can be nonzero only at blocks attaining the maximum spectral norm. For $\ell_\infty$, $\uu$ is supported on the largest-magnitude coordinates of $\thv$, with matching signs wherever $\uu$ is nonzero.

On strata: For $\ell_\infty$, this type is determined by the set $A$ of coordinates attaining the maximum absolute value and their signs. For the spectral norm, it is determined by the multiplicity $\rho$ of the top singular value; for the max-spectral norm, it also depends on which blocks attain the maximum. By ``on the generic stratum,'' we mean away from ties and other degeneracies: top singular values are simple, the $\ell_\infty$ maximum is unique, and the relevant active sets are locally constant. Unless stated otherwise, dimension counts are taken on this stratum.

On Assumption~\ref{ass:blockhom}: A first-layer bias is absorbed by appending a constant $1$ to the input, so that $[W_1\ \mathbf b_1]$ is a single block (and Muon is understood to act on it). Differentiability holds at almost every $\thv$ for ReLU networks, as used by \citet{haim2022}, and everywhere for the smooth homogeneous models covered by the guarantees of \citet{gronich2026}.

On the generalized Cauchy--Schwarz inequality: The three geometries considered here give familiar instances of this inequality: Cauchy--Schwarz for the $(\ell_2,\ell_2)$ pair, H\"older for $(\ell_\infty,\ell_1)$, and a bound obtained from von Neumann's trace inequality for the max-spectral norm and its dual, the sum-nuclear norm $\sum_k\|\uu_{W_k}\|_{\nuc}$.

\subsection{Other privacy attacks (from \S\ref{sec:related})}
\label{app:more-related}

Other privacy attacks differ in their goals and requirements. Membership inference determines whether an example belongs to the training set \citep{shokri2017}, model inversion can recover class-representative inputs \citep{fredrikson2015}, and gradient leakage reconstructs data from shared gradients \citep{zhu2019}. KKT-based reconstruction attacks use only the final weights of the trained model. Differentially private training \citep{abadi2016} provides a formal guarantee limiting the influence of individual training examples, potentially at a utility cost. Optimizer choice provides another way to affect reconstruction leakage, but does not provide the same guarantee.

\subsection{Implicit bias of gradient flow (from \S\ref{sec:notation})}
\label{app:implicit-bias}

\begin{theorem}[paraphrased version of \citet{Lyu2020Gradient}, \citet{jitelgarsky2020}]\label{thm:implicit_bias}
    Let $\Phi(\btheta; \bx)$ be a normalized homogeneous\footnote{A network $\Phi(\btheta; \bx)$ is called \emph{homogeneous} if there exists $c > 0$ such that for every $b > 0$, $\btheta$ and $\bx$, it holds that $\Phi(b \cdot \btheta; \bx) = b^c\Phi(\btheta; \bx)$ \cite{haim2022}.} ReLU neural network. 
    Consider minimizing the logistic ($z\mapsto \log(1+e^{-z})$) or the exponential ($z\mapsto e^{-z}$) loss using gradient flow (which is a continuous-time analog of gradient descent) over a binary classification set $\{(\bx_i, y_i)\}_{i=1}^n \subseteq \real^d \times \{-1 ,1\}$. Assume that there is a time $t_0$ where $L(\btheta(t_0)) < \frac{1}{n}$. Then, gradient flow converges in direction\footnote{We say that gradient flow \emph{converges in direction} to $\hat{\btheta}$ if $\lim_{t \rightarrow \infty}\frac{\btheta(t)}{\|\btheta(t)\|} = \frac{\hat{\btheta}}{\|\hat{\btheta}\|}$.} to a first order stationary point (KKT point \citep{cortes1995support}) of the maximum-margin problem~\eqref{eq:maximal_margin}.
\end{theorem}

A KKT point of \eqref{eq:maximal_margin} is characterized by the following set of conditions:
\begin{itemize}[noitemsep, topsep=2pt, parsep=0pt, partopsep=0pt, leftmargin=*]
    \item \textbf{Stationarity:} $\btheta = \sum_{i=1}^n \lambda_i \nabla_{\btheta}\left[y_i \Phi\left(\btheta ; \bx_i\right)\right]$
    \item \textbf{Primal feasibility:} $y_i \Phi\left(\btheta ; \bx_i\right) \geq 1, \quad \forall i \in [n]$
    \item \textbf{Dual feasibility:} $\lambda_i \geq 0, \quad \forall i \in [n].$
    \item \textbf{Comp. slackness:} $\lambda_i = 0 \text{ if } y_i \Phi\left(\btheta ; \bx_i\right) \neq 1, \quad \forall i \in [n].$
\end{itemize}

\subsection{Implicit-bias background (from \S\ref{sec:related})}
\label{app:more-related-ib}

The reconstruction attack relies on characterizing which interpolating solution an optimizer selects. For separable linear models, GD converges in direction to the $\ell_2$ maximum-margin solution \citep{soudry2018}. Related results establish convergence towards KKT directions of margin problems for homogeneous networks \citep{Lyu2020Gradient,jitelgarsky2020}, and connect steepest descent under other norms to the corresponding margin geometry \citep{gunasekar2018}; see \citet{vardi2023} for a survey. These results allow us to study optimizer choice through its associated margin geometry.

\end{document}